\documentclass[journal,final]{IEEEtran}
\usepackage{amsmath,amsfonts}
\usepackage{algorithmic}
\usepackage{algorithm}
\usepackage{array}
\usepackage{textcomp}
\usepackage{stfloats}
\usepackage{url}
\usepackage{verbatim}
\usepackage{graphicx}
\usepackage{cite}

\usepackage{multirow}
\usepackage[colorlinks,linkcolor=blue]{hyperref}
\usepackage{enumerate}
\usepackage{stmaryrd}
\usepackage{bigstrut}
\usepackage{ulem}
\usepackage{gensymb}
\usepackage[scaled=1.0]{helvet}
\usepackage{tabularx}
\usepackage{makecell}
\usepackage{subcaption}
\usepackage{threeparttable}

\newcommand\norm[1]{\left\lVert#1\right\rVert}

\begin{document}

\title{ALIKED: A Lighter Keypoint and Descriptor Extraction Network via Deformable Transformation}
%
%
%
\author{Xiaoming~Zhao,
    Xingming~Wu,
    Weihai~Chen*,~\IEEEmembership{Member,~IEEE},		
    Peter~C.~Y.~Chen,
    Qingsong~Xu,
    and~Zhengguo~Li,~\IEEEmembership{Fellow,~IEEE}
    \thanks{This work was supported by the National Nature Science Foundation of China under Grant No. 61620106012, the Key Research and Development Program of Zhejiang Province under Grant No. 2020C01109, and A*STAR of Singapore under Robotics Horizontal Technology Coordinating Office Project C221518005. (\textit{Corresponding authors: Weihai Chen.})}
    \thanks{Xiaoming Zhao, Xingming Wu, and Weihai Chen are with the School of Automation Science and Electrical Engineering, Beihang University, Beijing, 100191, China (e-mail: xmzhao@buaa.edu.cn, wxmbuaa@163.com, and whchen@buaa.edu.cn).}
    \thanks{Qingsong Xu is with Department of Electromechanical Engineering, Faculty of Science and Technology, University of Macau, Taipa, Macau, China (e-mail: qsxu@um.edu.mo).}
    \thanks{Peter C. Y. Chen is with the Department of Mechanical Engineering, National University of Singapore, Singapore (email: mpechenp@nus.edu.sg).}
    \thanks{Zhengguo Li is with SRO Department, Institute for Infocomm Research, Agency for Science, Technology and Research (A*STAR), 1 Fusionopolis Way, \#21-01, Connexis South Tower, Singapore 138632 (email: ezgli@i2r.a-star.edu.sg).}
}

%

\maketitle
	
\begin{abstract}
Image keypoints and descriptors play a crucial role in many visual measurement tasks. In recent years, deep neural networks have been widely used to improve the performance of keypoint and descriptor extraction. However, the conventional convolution operations do not provide the geometric invariance required for the descriptor. To address this issue, we propose the Sparse Deformable Descriptor Head (SDDH), which learns the deformable positions of supporting features for each keypoint and constructs deformable descriptors. Furthermore, SDDH extracts descriptors at sparse keypoints instead of a dense descriptor map, which enables efficient extraction of descriptors with strong expressiveness. In addition, we relax the neural reprojection error (NRE) loss from dense to sparse to train the extracted sparse descriptors. Experimental results show that the proposed network is both efficient and powerful in various visual measurement tasks, including image matching, 3D reconstruction, and visual relocalization.

\end{abstract}


%
\IEEEpeerreviewmaketitle

\section{Introduction}
	
Efficient and robust extraction of image keypoints and descriptors is critical for many resource-constrained visual measurement applications, such as simultaneous localization and mapping (SLAM) \cite{ogisalm2}, computational photography \cite{xu2022novel}, and visual place recognition \cite{yue2021automatic}. Early methods for keypoint detection and descriptor extraction relied on human heuristics \cite{sift,orb,rpi_surf}. However, these hand-crafted methods are not sufficiently efficient and robust. To address these issues, many data-driven approaches based on deep neural networks (DNNs) have emerged in recent years. Initially, DNNs were used to extract descriptors of image patches at predefined keypoints \cite{hardnet}. Subsequently, the mainstream approach became  the extraction of keypoints and descriptors with a single network \cite{superpoint,aslfeat,alike}, which can often extract more robust keypoints and discriminative descriptors than hand-crafted methods \cite{imw2020}. We refer to these methods as map-based methods because they estimate a score map and a descriptor map using two heads: the score map head (SMH) and the descriptor map head (DMH). Then they extract keypoints and descriptors from the score map and descriptor map, respectively. 

Existing map-based methods use fixed-size vanilla convolutions to encode images, which lack the geometric invariance that is essential for image matching performance. 
This problem can be alleviated by estimating the scale and orientation of the descriptors on the image \cite{lift,gift,lfnet,aslfeat}. However, the scale and orientation can only model affine transformations of the image features, not any geometric transformations of the image features. We observe that the deformable convolution network (DCN) \cite{dcn} can model any geometric transformation by adjusting the offset for each pixel in the convolution, thereby improving the representational capabilities of the descriptors. Unfortunately, DCN \cite{dcn} introduces additional computations when computing the dense descriptor map, which slows down the running speed. In order to improve the running speed when extracting deformable descriptors like DCN \cite{dcn}, we propose the Sparse Deformable Descriptor Head (SDDH). We studied existing map-based methods and found that the DMH has many redundant convolutions in areas without keypoints, leading to a high computational cost for descriptor extraction. The SDDH extracts deformable descriptors only at detected keypoints rather than on the entire dense feature map, making it more efficient since keypoints are usually fewer than dense image features. In addition, the SDDH estimates offsets at M sample locations, inspired by deformable image alignment \cite{chan2021understanding}, instead of using fixed-size convolutional grids like DCN \cite{dcn}. The estimated offsets are used to construct descriptors, and M can be any positive integer, making the SDDH more flexible and efficient in modeling deformable descriptors.

Afterward, we propose \textbf{A} \textbf{LI}ghter \textbf{K}eypoint and descriptor \textbf{E}xtraction network with \textbf{D}eformable transformation (ALIKED) for visual measurement using the SDDH. However, the SDDH extracts only  sparse descriptors, which means there are no descriptor maps for constructing neural reprojection error (NRE) loss \cite{nre, alike}. To address this challenge, we propose an elegant solution to relax the NRE loss from dense to sparse. Instead of constructing dense probability maps, we construct sparse probability vectors for sparse descriptors, and minimize the distance between the sparse matching and reprojection probability vectors. This approach not only overcomes the challenge of lacking dense descriptor maps but also reduces redundant computations during network training, resulting in significant memory savings on the graphics processing unit (GPU).

Overall, the main contributions of this paper are as follows: 
\begin{itemize}
\item We propose the SDDH for efficient extraction of deformable descriptors, greatly reducing redundant computations and allowing the modeling of any geometric transformation.
\item We develop the ALIKED network for visual measurement using the SDDH, which includes an elegant solution to relax the NRE loss from dense to sparse, allowing sparse descriptor training for NRE loss and reducing redundant computations during network training.
\item Experimental results demonstrate that the ALIKED network achieves excellent performance in various visual measurement tasks, including image matching, 3D reconstruction, and visual localization.
\end{itemize}

For quick reference, Table \ref{tab_abbreviation} lists the most commonly used abbreviations in this paper. The rest of the paper is organized as follows: Section \ref{sec_related} reviews the deep learning-based keypoint and descriptor extraction methods for visual measurements. Section \ref{sec_network} first introduces the overall network architecture, and then section \ref{sec_def} and section \ref{sec_loss} present the SDDH and the loss functions, respectively. Section \ref{sec_exp} presents the comparisons with state-of-the-art (SOTA) methods and ablation studies, and we conclude our work in Section \ref{sec_con}.

\section{Related Works}
\label{sec_related}
In this section, we review the geometric modeling for descriptor extraction, the deep learning-based keypoint extraction network used in visual measurement systems, and the use of deformable convolutions in neural networks.

\subsection{Geometric Invariant Descriptor Extraction}
In hand-crafted methods, the geometric invariance of descriptors is typically defined in two aspects: scale invariance and orientation invariance. For example, SIFT \cite{sift} estimates the scale of each detected keypoint in the scale space and computes the keypoint orientation based on the histogram of image gradients. SIFT also extracts image patches using the estimated scale and orientation and constructs descriptors based on these image patches. On the other hand, ORB \cite{orb} features extract orientations only for keypoints at the center of mass for efficiency, then the image patches are rotated to achieve orientation invariance.

In terms of learning methods, there are two main approaches: patch-based descriptor extraction methods and joint keypoint and descriptor learning methods. Patch-based methods \cite{matchnet, tfeat, l2net, hardnet, sosnet} as well as most joint keypoint and descriptor learning methods \cite{superpoint, disk, d2net} rely on data augmentation to achieve scale and orientation invariance. Some of the joint learning methods explicitly model the orientation and scale for keypoints. For example, LIFT \cite{lift} mimics the SIFT \cite{sift} by detecting keypoints, estimating their orientations, and extracting descriptors with different neural networks. It estimates the orientation and scale for keypoints with a neural network and applies the transformation to the obtained features for orientation- and scale-invariant descriptor extraction. Similarly, AffNet \cite{mishkin2018repeatability}, UCN \cite{ucn}, and LF-Net \cite{lfnet} estimate affine parameters and apply affine transformations on image features using Spatial Transformer Networks (STN) \cite{stn} to extract affine invariant descriptors. GIFT \cite{gift} first generates groups of images with different scales and orientations, and then extracts features from these images to produce scale and orientation invariant descriptors. HDD-Net \cite{hddnet} suggests rotating the convolution kernels instead of the features to extract rotation invariant descriptors.

In the above methods, the geometric transformation is predefined as an affine transformation. Inspired by the ASLFeat \cite{aslfeat}, the proposed network ALIKED also employs DCN to extract geometric invariant features. Moreover, based on the deformable philosophy of DCN \cite{dcn}, we design the SDDH module to efficiently extract geometric invariant descriptors.

\begin{table}[t]
	\centering
	\caption{The most commonly used abbreviations in this paper.}
	\begin{tabular}{cc}
		\hline
		\textbf{Abbreviation} & \multicolumn{1}{c}{\textbf{Explanation}} \\
		\hline
		\textbf{ALIKED} & The proposed method. \\
		\textbf{CNN} & Convolutional Neural Network. \\
		\textbf{DCN} & Deformable Convolution Network. \\
		\textbf{DKD} & Differentiable Keypoint Detection. \\
		\textbf{DMH} & Descriptor Map Head. \\
		\textbf{DNNs} & Deep Neural Networks. \\
		\textbf{FPS} & Frames Per Second. \\
		\textbf{GFLOPs} & Giga FLoating-point OPerations. \\
		\textbf{GPU} & Graphics Processing Unit. \\
		\textbf{mAA} & mean Average Accuracy. \\
		\textbf{MHA} & Mean Homography Accuracy. \\
		\textbf{MMA} & Mean Matching Accuracy. \\
		\textbf{mNN} & mutual Nearest Neighbor. \\
		\textbf{MP} & Million numbers of Parameters. \\
		\textbf{MS} & Matching Score. \\
		\textbf{NMS} & Non-Maximum Suppression. \\
		\textbf{NRE} & Neural Reprojection Error. \\
		\textbf{NSGD} & Normalized Symmetric Geometric Distance. \\
		\textbf{PPC} & Performance Per Cost. \\
		\textbf{Rep} & Repeatability. \\
		\textbf{SDDH} & Sparse Deformable Descriptor Head. \\
		\textbf{SLAM} & Simultaneous Localization And Mapping.  \\
		\textbf{SMH} & Score Map Head. \\
		\textbf{SOTA} & State-Of-The-Art. \\
		\textbf{TL } & Track Length. \\
		\hline
	\end{tabular}%
	\label{tab_abbreviation}%
\end{table}%

\subsection{Joint Keypoint and Descriptor Learning}

Many studies propose to jointly estimate the score map and the descriptor map, detect keypoints from the score map, and sample descriptors from the descriptor map. The SuperPoint \cite{superpoint} proposes a lightweight network that is trained on homography image pairs generated from Homographic Adaptation. R2D2 \cite{r2d2} computes the repeatability and reliability maps for keypoint detection, and it trains the descriptors with AP loss. Suwichaya recently added a low-level feature LLF detector to the R2D2 to improve keypoint accuracy \cite{suwanwimolkulLearningLowlevelFeature2021a}. DISK \cite{disk} uses reinforcement learning to train the score map and descriptor map. ALIKE \cite{alike} has a differentiable keypoint detection module for accurate keypoint training and has the lightest network, thereby allowing its application in real-time visual measurement applications. D2-Net \cite{d2net} does not estimate the score map with the network, but rather detects keypoints with channel and spatial maxima on the feature map. However, because it extracts keypoints from a low-resolution feature map, D2-Net \cite{d2net} lacks accuracy in keypoint localization. ASLFeat \cite{aslfeat} uses a multi-level feature to detect the keypoint and models the local shape with deformable convolutions to improve localization accuracy and descriptors. D2D \cite{d2d}, inspired by D2-Net \cite{d2net}, detects keypoints on a feature map using a descriptor map and absolute and relative saliency. Rao et al. proposed the hierarchical view consistency for general feature descriptors\cite{hvc} for visual measurements.

Despite significant advances in joint keypoint and descriptor learning, their complexity remains the primary obstacle to visual measurement applications. Most of these methods extract dense but expensive descriptor maps to improve matching performance, which is computationally expensive. To address this problem, we extract descriptors on deformable local features for each sparse keypoint instead of dense descriptor maps. As a result, we improve the lightweight ALIKE \cite{alike} and propose ALIKED with deformable features and descriptor extraction using computational budget savings.

\begin{figure*}[!t]
\centering
\includegraphics[]{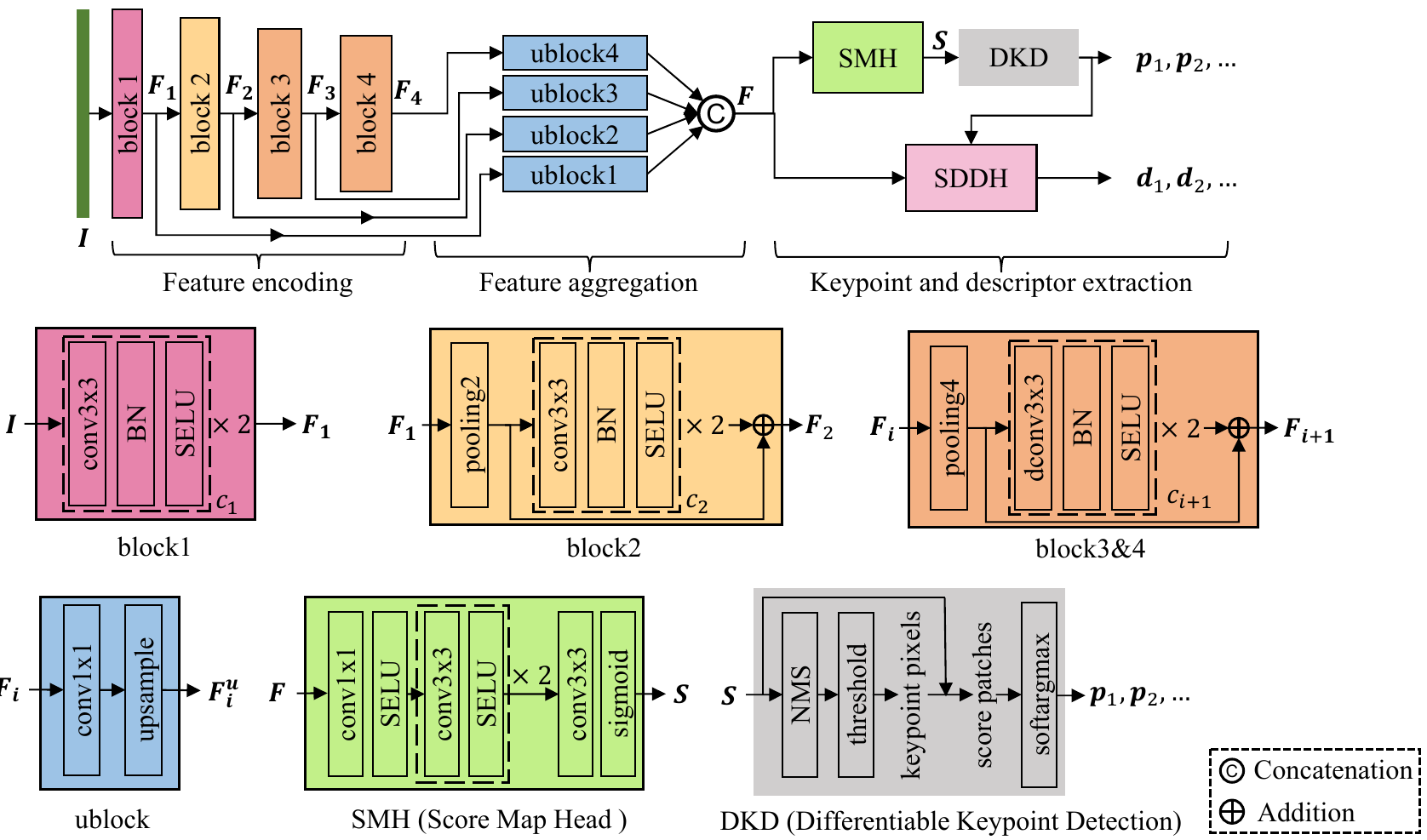}
\caption{The network architecture of ALIKED. The input image $\boldsymbol{I}$ is initially encoded into multi-scale features $\{\boldsymbol{F}_1,\boldsymbol{F}_2,\boldsymbol{F}_3,\boldsymbol{F}_4\}$ with encoding block1 to block4, and the number of channels of $\boldsymbol{F}_i$ is $c_i$ (detailed settings are listed in Table \ref{tab_size}). Then, the multi-scale features are aggregated with upsample blocks (ublock4 to ublock1), and the output features $\boldsymbol{F}_i^u$ are concatenated to obtain the final image feature $\boldsymbol{F}$. The Score Map Head (SMH) extracts the score map $\boldsymbol{S}$ with $\boldsymbol{F}$ followed by a Differentiable Keypoint Detection (DKD) module \cite{alike} to detect the keypoints $\{\boldsymbol{p}_1,\boldsymbol{p}_2,...\}$. The SDDH (as shown in Fig. \ref{fig_SDDH} and is proposed in Section \ref{sec_def}) then efficiently extracts deformable invariant descriptors at the detected keypoints. ``BN'', ``poolingN'', and ``DCN3x3'' denote batch normalization, $N\times N$ average pooling, and $3\times3$ deformable convolution \cite{dcn}, respectively.}
\label{fig_net}
\end{figure*}  

\subsection{Deformable Convolutions in Neural Networks}
Regular CNNs have fixed convolutional kernels, which limits the ability to exploit long-range information. To address this issue, the deformable convolution introduces learnable offsets for convolution kernels \cite{dcnv1,dcn}. This approach has been shown to be effective in high-level tasks such as object detection \cite{bertasius2018object}, semantic segmentation \cite{dcnv1}, action recognition \cite{zhao2018trajectory}, and human pose estimation \cite{sun2018integral}. It has also been widely used in low-level tasks, including video super-resolution \cite{tian2020tdan}, high dynamic range images \cite{liu_adnet_2021}, and video frame interpolation \cite{shi_video_2022}. The DCN has also been used in ASLFeat for descriptor extraction \cite{aslfeat}. However, ASLFeat only uses the DCN to compute dense features, while the proposed SDDH is specifically designed for efficient sparse descriptor extraction.

\begin{table}[t]
	\centering
	\caption{The network configurations. ``$c_{i}$'' denotes the channel numbers of features in the $i$-th block, and ``$dim$'' is the dimension of the output descriptor.}
	\begin{threeparttable}
		\begin{tabular*}{\linewidth}{@{}@{\extracolsep{\fill}}lrrrrr@{}}
			\hline
			\textbf{Models} & $c_1$ & $c_2$ & $c_3$ & $c_4$ & $dim$  \bigstrut \\ \hline
			Tiny (-T)   &     8 &    16 &    32 &    64 &    64  \bigstrut[t]\\
			Normal (-N) &    16 &    32 &    64 &   128 &   128 \\
			Large (-L)  &    32 &    64 &   128 &   128 &   128\tnote{*} \\ \hline
		\end{tabular*}%
		\begin{tablenotes}
			\item[*] In the large model, the descriptor head has two $1\times1$ conv layers \cite{alike}.
		\end{tablenotes}	
	\end{threeparttable}
	\label{tab_size}%
\end{table}%

In recent years, the vision transformer \cite{vit} has received considerable attention for its impressive performance. However, this model inherits the multi-head self-attention mechanism \cite{vaswani_attention_2017}, resulting in high computational burden during image feature extraction. To address this problem, the deformable DETR model proposes the use of DCN to attend to a small set of sample positions \cite{zhu_deformable_2020}. More recently, InternImage introduced DCNv3 \cite{wang_internimage_2022}, which not only reduces the computational burden of vision transformers, but also achieves state-of-the-art performance on basic vision tasks. Our approach follows a similar philosophy to that of InternImage \cite{wang_internimage_2022}. We compute descriptors only on sparse keypoints, thereby improving both the computational efficiency and the performance.

\section{Network Architecture of ALIKED}
\label{sec_network}

In this section, we will first introduce the overall architecture of ALIKED. As shown in Fig. \ref{fig_net}, ALIKED  consists of three components: feature encoding, feature aggregation, and keypoint and descriptor extraction. Then, in next section, we will present the inspirations and design considerations of SDDH in ALIKED.

\subsection{Feature Encoding}
\label{sec_fea}
	
The feature encoder transforms the input image $\boldsymbol I \in \mathbb{R}^{H\times W\times 3}$ into multi-scale features ${\boldsymbol{F}_1,\boldsymbol{F}_2,\boldsymbol{F}_3,\boldsymbol{F}_4}$ using four encoding blocks, each with a channel range from $c_1$ to $c_4$ (detailed settings are listed in Table \ref{tab_size}). The first block, as shown in Fig. \ref{fig_net}, consists of two convolutions that extract low-level image features $\boldsymbol{F}_1$. To cover larger receptive fields and increase computational efficiency, the second block uses $2\times2$ average pooling to downsample $\boldsymbol{F}_1$. The third and fourth blocks first downsample the features using $4\times4$ average pooling and then extract the image features (Section \ref{sec_def}) using the residual block with $3\times3$ DCNs \cite{dcn}. To improve convergence, the ALIKED model uses SELU \cite{selu} activation functions instead of ReLU \cite{relu}.

\subsection{Feature Aggregation}

The feature aggregation part is responsible for aggregating multi-scale features $\{\boldsymbol{F}_1,\boldsymbol{F}_2,\boldsymbol{F}_3,\boldsymbol{F}_4\}$ for both localization and representation abilities. As shown in Fig. \ref{fig_net}, four ublocks are used to aggregate these features. Each ublock consists of a $1\times1$ convolution and an upsample layer to align the dimensions and resolutions of the multi-scale features. By concatenating these aligned features $\{\boldsymbol{F}_1^u,\boldsymbol{F}_2^u,\boldsymbol{F}_3^u,\boldsymbol{F}_4^u\}$, we obtain the aggregated feature $\boldsymbol{F}$ for keypoint and descriptor extraction.

\subsection{Differentiable Keypoint Detection}
For keypoint detection, the Score Map Head (SMH) estimates the score map $\boldsymbol{S} \in \mathbb{R}^{H\times W}$ using the aggregated feature $\boldsymbol{F}$. As shown in Fig. \ref{fig_net}, the SMH first uses a $1\times1$ convolution layer to reduce the feature channels to eight, followed by two $3\times3$ convolution layers for feature encoding. Finally, a $3\times3$ convolution layer and a sigmoid activation layer are used to obtain the score map $\boldsymbol{S}$.

ALIKED uses Differentiable Keypoint Detection (DKD) \cite{alike} to detect trainable differentiable keypoints. As shown in Fig. \ref{fig_net}, the DKD module first applies non-maximum suppression (NMS) to the score map $\boldsymbol{S}$ to identify local maxima. The pixel-level keypoints are then determined by setting a threshold for the local maximum scores. The DKD module further improves the accuracy of the pixel-level keypoints by refining their positions with softargmax on the local patches, thus extracting differentiable subpixel keypoints ($\boldsymbol{P}={\boldsymbol{p}_1,\boldsymbol{p}_2,...}$). By using these differentiable keypoints $\boldsymbol{P}$, we can directly optimize the reprojection error of the corresponding keypoints between images (as described in Section \ref{sec_kpt_rep_loss}) to train the score map.

\section{Sparse Deformable Descriptor Head}
\label{sec_def}
In this section, we present the Sparse Deformable Descriptor Head in the ALIKED, as shown in Fig. \ref{fig_net}.

\subsection{Deformable Invariant Descriptor Modeling}
Existing hand-craft methods \cite{sift} model the geometric invariance of descriptors with the affine transformation on a local image patch as
\begin{equation}
	\begin{bmatrix}
		x'\\
		y'\\
	   1
	   \end{bmatrix}=\begin{bmatrix}
		\boldsymbol{A}  & \boldsymbol{b}  \\
		\boldsymbol{0}  & 1
	   \end{bmatrix}\begin{bmatrix}
		x\\
		y\\
	   1
	   \end{bmatrix},
	   \label{equ_aff}
\end{equation}
where $[x,y,1]^T$ and $[x',y',1]^T$ are the homogeneous coordinates before and after the transformation, respectively. $\boldsymbol{A}\in \mathbb{R}^{2\times2}$ and $\boldsymbol{b}\in \mathbb{R}^{2}$ denote the affine matrix and the bias, respectively. Unfortunately, conventional convolutions cannot directly preserve the affine invariance. To address this issue, some methods explicitly rotate and scale the images \cite{gift} or convolution kernels \cite{hddnet} with predefined degrees and scales. However, the local shape of the image keypoints can be much more complex than the affine transformation. Therefore, we model the geometric transformation as the following deformable transformation:
\begin{equation}
	\begin{bmatrix}
		x'\\
		y'\\
	   \end{bmatrix}=\begin{bmatrix}
		x\\
		y\\
	   \end{bmatrix}+\begin{bmatrix}
		\Delta x\\
		\Delta y\\
	   \end{bmatrix},
	   \label{equ_def}
\end{equation}
where $[\Delta x,\Delta y]^T$ is the offset for each pixel around the keypoint. Unlike the affine transformation \eqref{equ_aff}, which has six degrees of freedom for a local image patch, the deformable transformation \eqref{equ_def} has a degree of freedom equal to the number of pixels. As a result, the deformable transformation can provide general geometric invariance for keypoint descriptors.

\subsection{The Design of Sparse Deformable Descriptor Head}
\label{sec_SDDH}
Most learning-based keypoint and descriptor extraction methods \cite{superpoint, r2d2, alike} first encode the image with a convolutional network into a dense descriptor map and then sample the descriptors from the dense descriptor map. However, extracting the dense descriptor map can be very inefficient. Based on the requirement of efficient and geometrically invariant descriptor extraction, we design the SDDH.

\begin{figure*}[!t]
    \centering
    \includegraphics[]{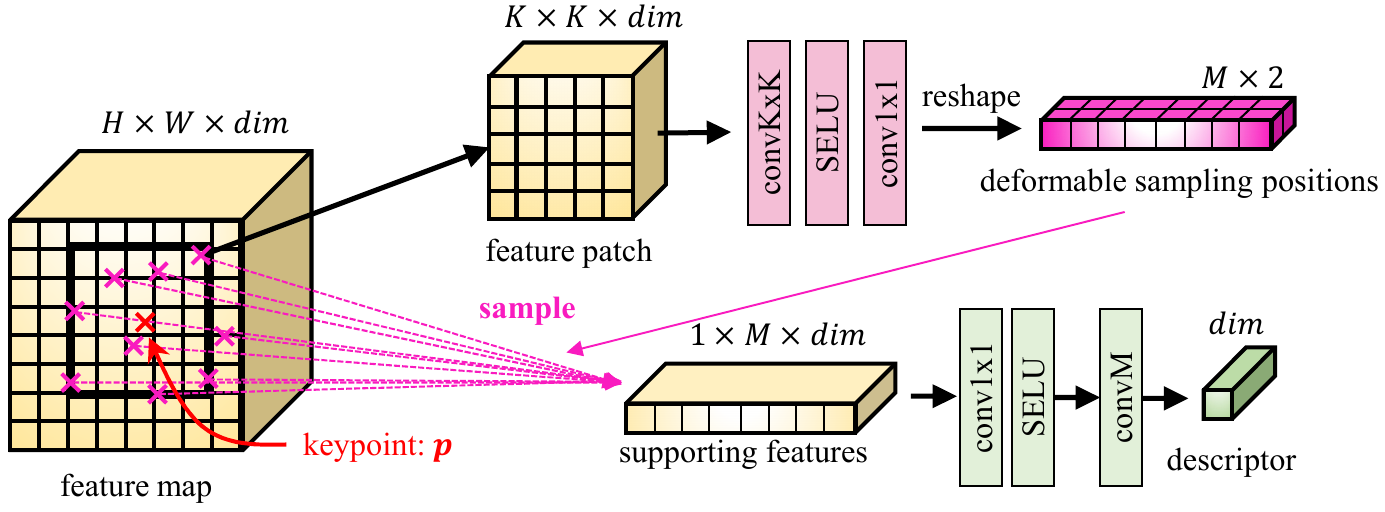}
    \caption{The SDDH estimates $M$ deformable sample positions on $K \times K$ keypoint feature patches ($K=5$ in this example), samples $M$ supporting features on the feature map based on the deformable sample positions, encodes the supporting features, and aggregates them with convM for descriptor extraction.}
    \label{fig_SDDH}
\end{figure*}

\begin{figure}[!t]
    \centering
    \includegraphics[width=0.95\linewidth]{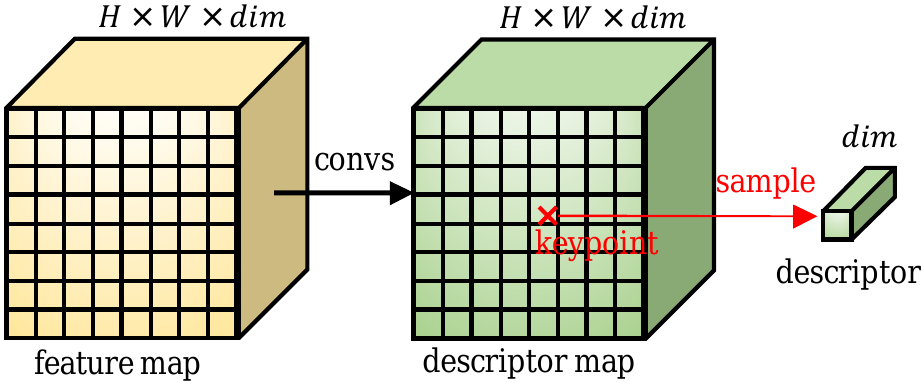}
    \caption{The DMH extracts a dense descriptor map by performing convolutions on the dense feature map $\boldsymbol F$. The descriptors are then sampled from the dense descriptor map at keypoint locations.}
    \label{fig_dmh}
\end{figure}

\subsubsection{DCN Revisited}
The DCN \cite{dcn} estimates sample offsets in the convolution and can be used to extract deformable invariant features. Consider a point $\boldsymbol p$ on the feature map $\boldsymbol F \in \mathbb{R}^{H\times H\times dim}$. Let $\boldsymbol{p}_i\in \mathbb{R}^{2}$ denote the i-th sample position on local feature patches for $K\times K$ convolution (for example, if $K=3$, $\boldsymbol{p}_i\in\{(-1,-1),(-1,0),...,(1,1)\}$). The deformable convolution of feature $\boldsymbol F$ is given as
\begin{equation}
    \boldsymbol{F}'(\boldsymbol{p})=\sum_{i=1}^{K^2} w(\boldsymbol{p}_i)\boldsymbol{F}(\boldsymbol{p}+\boldsymbol{p}_i+\Delta \boldsymbol{p}_i),
    \label{equ_dcn}
\end{equation}
where $\boldsymbol{F}'(\boldsymbol{p}) \in \mathbb{R}^{dim}$ denotes the output feature, $\boldsymbol{w}\in \mathbb{R}^{K\times K}$ is the convolution weights, and $\Delta\boldsymbol{p}_i$ is the offset for the sample position $\boldsymbol{p}_i$. The DCN \cite{dcn} uses basic convolutions to estimate offsets and extract features, and is therefore suitable for geometrically related features extraction by stacking multiple layers. Therefore, we use DCN \cite{dcn} in block3 and block4 of ALIKED (Section \ref{sec_fea}).

\subsubsection{Descriptor Map Head Revisited}
Existing methods \cite{superpoint, r2d2, alike} use convolutional layers to encode a dense feature map $\boldsymbol F$ into a dense descriptor map, from which descriptors are sampled. We refer to this module as the DMH, as shown in Fig. \ref{fig_dmh}. However, convolution on an image resolution feature map is computationally expensive, so some existing methods to downsample the feature map \cite{superpoint} or use lightweight operations \cite{r2d2, alike} to extract the dense descriptor map. Consequently, the representational ability of the descriptors is limited. In contrast, we argue that the dense descriptor map is unnecessary since only sparse descriptors corresponding to keypoints are needed. By eliminating the dense descriptor map, the computational effort can be reduced, allowing the extraction of more powerful descriptors while achieving computational savings.

\subsubsection{Sparse Deformable Descriptor Head}
Although DCN \cite{dcn} can extract deformable invariant features, it cannot efficiently and effectively extract sparse descriptors because it performs dense and simple convolutions on the feature map. To address this issue, we propose the SDDH for efficient extraction of sparse deformable descriptors, which builds on the philosophy of DCN \cite{dcn}, as shown in Fig. \ref{fig_SDDH}. For a given keypoint $\boldsymbol{p}\in \mathbb{R}^{2}$, the SDDH first extracts a feature patch $\boldsymbol{F}_{K\times K}$ of size $K\times K$, centered at $\boldsymbol{p}$ ($K=5$ in Fig. \ref{fig_SDDH}). Then  it estimates the deformable sample position $\boldsymbol{p}^s \in \mathbb{R}^{M\times 2}$ for the keypoint $\boldsymbol{p}$:
\begin{equation}
    \boldsymbol{p}^s = \operatorname{conv1x1}(\operatorname{SELU}(\operatorname{convKxK}(\boldsymbol{F}_{K\times K}))),
\end{equation}
where the number of output channels for both $\operatorname{convKxK}$ and $\operatorname{conv1x1}$ is $2M$, but only $\operatorname{convKxK}$ has no padding. Using the deformable sample position $\boldsymbol{p}^s \in \mathbb{R}^{M\times 2}$, the SDDH samples the supporting features on the feature map using bilinear sampling. The descriptor $\boldsymbol{d}\in \mathbb{R}^{dim}$ is then obtained as 
\begin{equation}
    \boldsymbol{d} = \sum_{i=1}^{M} w_M(\boldsymbol{p}_i) \Phi(\boldsymbol{F}(\boldsymbol{p}+\boldsymbol{p}_i^s)),	
    \label{equ_d}	
\end{equation}
where $\Phi(\boldsymbol{x})=\operatorname{SELU}(\operatorname{conv1x1}(\boldsymbol{x}))$. And $\boldsymbol{w}_M$ denotes the weights for convM (Fig. \ref{fig_SDDH}). The convM is the weighted summation operation \eqref{equ_d}, which is similar to the convolution except that it calculates over $M$ flexible positions instead of $K\times K$ fixed positions.

The SDDH differs from DCN \cite{dcn} in the following aspects: 
\begin{itemize}
\item The DCN \cite{dcn} uses deformable convolutions on a dense feature map, while the SDDH extracts deformable features only for sparse keypoints. As a result, the SDDH can significantly reduce the computational cost, since the number of keypoints is usually much smaller than the number of image pixels.
\item The DCN \cite{dcn} estimates $K\times K$ offsets for the convolution (as shown in equation \eqref{equ_dcn}), while the SDDH estimates $M$ deformable sample positions. Unlike the DCN, which is limited by the requirement that the sampled positions are the fixed $K\times K$ grid, the SDDH can be used with any positive integer value of $M$. This makes the SDDH more flexible than the DCN in terms of both performance and computational efficiency.
\item The DCN \cite{dcn} typically uses a simple network and is stacked with multilayers for feature extraction. In contrast, the SDDH uses a more sophisticated network to estimate the positions of deformable samples and extract deformable descriptors directly.
\end{itemize}

\begin{table*}[htbp]
  \centering
  \caption{The efficiency comparison between DMH and SDDH. The details are discussed in Section \ref{sec_complexity}.}
  \setlength{\tabcolsep}{1.2mm}
	{
    \begin{tabular*}{\linewidth}{@{}@{\extracolsep{\fill}}|c|r|r|r|r|r|@{}}
    \hline
    \textbf{Method} & \multicolumn{2}{c|}{\textbf{DMH}} & \multicolumn{3}{c|}{\textbf{SDDH}} \bigstrut\\
    \hline
    \textbf{Stage} & \multicolumn{1}{c|}{convolutions} & \multicolumn{1}{c|}{descriptor sample} & \multicolumn{1}{c|}{sample position estimation} & \multicolumn{1}{c|}{feature sample} & \multicolumn{1}{c|}{descriptor extraction} \bigstrut\\
    \hline
    \textbf{Theoretical complexity} & $HWC^2(K^2+1) $ & $4NC$  & $2NM(K^2C+2M)$ & $4NMC$ & $2NMC^2$  \bigstrut\\
    \hline
    \textbf{Complexity (K=5, N=5000)} & 130.86G & 2.56M   & 812.50M & 64.00M & 4096.00M  \bigstrut\\
    \hline
    \textbf{Complexity (K=3, N=1000)} & 50.33G & 2.56M   & 21.06M & 4.61M  & 294.91M  \bigstrut\\
    \hline
    \textbf{Running time (K=5, N=5000)} & 50.79ms & 1.06ms  & 3.42ms & 1.56ms & 2.62ms  \bigstrut\\
    \hline
    \textbf{Running time (K=3, N=1000)} & 14.42ms & 0.28ms  & 0.58ms & 0.30ms & 0.37ms  \bigstrut\\
    \hline
    \end{tabular*}%
    }
  \label{tab_complexity}%
\end{table*}%

\subsection{Efficiency Comparison between DMH and SDDH}
\label{sec_complexity}
We demonstrate the efficiency of the SDDH by comparing its computational operations with those of the DMH on an $H\times W\times C$ feature map with $N$ keypoints. For the DMH with $\operatorname{convKxK}(\operatorname{SELU}(\operatorname{conv1x1}(\boldsymbol{x})))$ and a convolution kernel size of $K=5$, the equivalent SDDH is the one with $M=K^2$. Table \ref{tab_complexity} shows the theoretical and typical computational complexity and running time for the DMH and the SDDH.

For the DMH, the theoretical computational operations of $\operatorname{conv5x5}$ and $\operatorname{conv1x1}$ are $HWK^2C^2$ and $HWC^2$, respectively, for a total of $HWC^2(K^2+1)$. For bilinear sampling, the operations for sampling a descriptor from the descriptor map of channel $C$ are $4C$, and the operations for sampling $N$ descriptors are $4NC$. For the SDDH, the theoretical computational operations for estimating the deformable sample position of a keypoint are $(K^2C\times2M+2M\times2M)=2M(K^2C+2M)$, and the total operations for $N$ keypoints are $2NM(K^2C+2M)$. Sampling $M$ deformable features for $N$ keypoints would require $4NMC$ operations. At the descriptor extraction stage, the operations of $\operatorname{conv1x1}$ and $\operatorname{convM}$ are both $NMC^2$ for a total of $2NMC^2$.

To provide a more intuitive comparison, we report the typical complexity and running time of two configurations (K=5, N=5000) and (K=3, N=1000) when the feature map is $480\times 640\times 128$ in Table \ref{tab_complexity}. The running times were evaluated on a mid-end GPU, specifically the NVIDIA GeForce RTX 2060. In both cases, the DMH spends significant computational resources on convolutions to extract the dense descriptor map. In contrast, the SDDH only performs computations on sparse keypoint patches, making it much more efficient than the DMH. The superiority of the SDDH is more evident for smaller patch sizes and fewer keypoints.

\section{The Loss Functions}
\label{sec_loss}

In this section, we introduce the loss function used to train ALIKED. To supervise the keypoints, we adopt the reprojection loss and the dispersity peak loss originally proposed in ALIKE \cite{alike}. Since there is no dense descriptor map available to compute the matching probability map, we propose to relax the Neural Reprojection Error (NRE) loss \cite{nre, alike} from dense to sparse. In addition, we introduce a reliable loss based on sparse descriptor similarity.

Considering an image pair $(\mathcal{I}_A, \mathcal{I}_B)$, the network extracts the score maps $\boldsymbol{S}_A$ and $\boldsymbol{S}_B$ from an image pair $(\mathcal{I}_A, \mathcal{I}_B)$, from which the the DKD module then detects the keypoints $\boldsymbol{P}_A \in \mathbb{R}^{N_A\times 2}$ and $\boldsymbol{P}_B \in \mathbb{R}^{N_B\times 2}$, respectively. Corresponding descriptors for $\boldsymbol{P}_A$ and $\boldsymbol{P}_B$ are denoted by $\boldsymbol{D}_A \in \mathbb{R}^{N_A\times dim}$ and $\boldsymbol{D}_B \in \mathbb{R}^{N_B\times dim}$, respectively. We define all loss functions as follows:

\subsection{Reprojection Loss}
\label{sec_kpt_rep_loss}

Since the keypoints extracted from the ALIKED are differentiable, we can directly train the position of the keypoints using the reprojection distance \cite{alike}. First, for a keypoint $\boldsymbol{p}_A$ in image $\mathcal{I}_A$, we warp it to image $\mathcal{I}_B$ using 3D perspective projection:
\begin{equation}
    \boldsymbol{p}_{AB} = \pi( d_A\boldsymbol{R}_{AB}\pi^{-1}(\boldsymbol{p}_A) + \boldsymbol{t}_{AB}),
\end{equation}
where $\boldsymbol{R}_{AB}$ and $\boldsymbol{t}_{AB}$ denote the rotation and translation matrix from $\mathcal{I}_A$ to $\mathcal{I}_B$, respectively. $d_A$ represents the depth of $\boldsymbol{p}_A$, and $\pi(\boldsymbol{P})$ is the process to project a 3D point $\boldsymbol{P}=[X,Y,Z]^T$ to the image plane.
In $\mathcal{I}_B$, we search for the nearest keypoint $\boldsymbol{p}_B$ to $\boldsymbol{p}_{AB}$, and their distance must be less than $th_{gt}$ pixels. This keypoint $\boldsymbol{p}_B$ is considered to be the matching keypoint of $\boldsymbol{p}_A$. Similarly, we also project the $\boldsymbol{p}_B$ back to $\mathcal{I}_A$ to obtain $\boldsymbol{p}_{BA}$.
The reprojection loss of $(\boldsymbol{p}_A,\boldsymbol{p}_B)$ is defined as 
\begin{equation}
	\mathfrak{L}_{rp}(\boldsymbol{p}_A,\boldsymbol{p}_B) = \frac{1}{2}(\left\lVert \boldsymbol{p}_A-\boldsymbol{p}_{BA} \right\rVert + \left\lVert \boldsymbol{p}_B-\boldsymbol{p}_{AB} \right\rVert),
\end{equation}
The overall reprojection loss $\mathfrak{L}_{rp}$ is then calculated as the average reprojection loss of all matching keypoints in both images.

\subsection{Dispersity Peak Loss}

The dispersity peak loss aims to maximize the scores precisely at the keypoint \cite{alike}. In the DKD module, assuming a window size of $W$, we can obtain a $W\times W$ score patch $\boldsymbol{S}_p$ on the score map $\boldsymbol{S}$ corresponding to a keypoint $\boldsymbol{p}$. The dispersity peak loss is defined as the product of the softmax score of the patch and the distance between each coordinate $\boldsymbol{c}$ in $\boldsymbol{S}_p$ and the keypoint:
\begin{equation}
\mathfrak{L}_{pk}(\boldsymbol{p}) = \operatorname{mean}(\operatorname{softmax}(\boldsymbol{s}_p)\cdot \norm {\boldsymbol{p} - \boldsymbol{c}}),
\end{equation}
where $\cdot$ denotes the dot product, $\boldsymbol{s}_p \in \boldsymbol{S}_p$ represents the score in the patch, and the $\operatorname{softmax}$ function is defined as $\operatorname{softmax} (\boldsymbol{x}) = \exp (\boldsymbol{x}) / \sum_i \exp (\boldsymbol{x_i})$.
To obtain the overall dispersity peak loss, we calculate the average dispersity peak loss for all keypoints in both images.

\subsection{Sparse Neural Reprojection Error Loss}

Theoretically, matching keypoints in different images should have identical descriptors. Conversely, the descriptors for non-matching keypoints should be distinct. One method to achieve this property is to use the dense NRE loss, which uses the cross-entropy loss to minimize the difference between the reprojection probability map and the matching probability map \cite{alike}. However, the SDDH only produces sparse descriptors, and a descriptor map is not available for generating matching probability maps. To overcome this limitation, we relax the probability map from dense to sparse.

Let the descriptor of $\boldsymbol{p}_A$ be $\boldsymbol{d}_A$. Without the dense descriptor map, we can still define the reprojection probability $q_r(\boldsymbol{p}_A, \boldsymbol{P}_B)$ for $\boldsymbol{p}_A$ with respect to $\boldsymbol{P}_B$ as a binary vector, where a true element indicates the matching keypoint of $\boldsymbol{p}_A$ in $\boldsymbol{P}_B$. Similarly, we can also construct a matching similarity vector of $\boldsymbol{d}_A$ and $\boldsymbol{D}_B$ as follows:
\begin{equation}
	\label{equ_sim}
	\operatorname{sim}(\boldsymbol{d}_A, \boldsymbol{D}_B) = \boldsymbol{D}_B \boldsymbol{d}_A,
\end{equation}
where $\boldsymbol{D}_B\in \mathbb{R}^{N_B\times dim}$ is the descriptors for all keypoints in image $\mathcal{I}_B$. Then the matching probability vector is
\begin{equation}
	\label{equ_qm}
	q_m(\boldsymbol{d}_A, \boldsymbol{D}_B) = \operatorname{softmax}((\operatorname{sim}(\boldsymbol{d}_A, \boldsymbol{D}_B)-1)/t_{des}),
\end{equation}
where $t_{des}$ controls the sharpness of the matching probability. Then we can define the sparse NRE loss as the cross-entropy (CE) between the reprojection probability vector $q_r(\boldsymbol{p}_A, \boldsymbol{P}_B)$ and the matching probability vector $q_m(\boldsymbol{d}_A, \boldsymbol{D}_B)$:
\begin{equation}
	\begin{aligned}
		\mathfrak{L}_{ds}(\boldsymbol{p}_A, \mathcal{I}_B) & = CE\left(q_r(\boldsymbol{p}_A, \boldsymbol{P}_B) \| q_m(\boldsymbol{d}_A, \boldsymbol{D}_B) \right) &  \\
		& = - \ln \left( q_m(\boldsymbol{d}_A, \boldsymbol{d}_B) \right),                                      &
	\end{aligned}
\end{equation}
where $\boldsymbol{d}_B$ is the descriptor of the matching keypoint in $\mathcal{I}_B$. We can obtain sparse NRE loss $\mathfrak{L}_{ds}(\boldsymbol{p}_B, \mathcal{I}_A)$ for keypoint $\boldsymbol{p}_B$ in the same manner. The overall sparse NRE loss $\mathfrak{L}_{ds}$ is the average sparse NRE loss for all descriptors in both images.

\subsection{Reliable Loss}

The score map shows the probability of a pixel being a keypoint, but reliability should also be considered, as suggested in R2D2 \cite{r2d2}. Areas of low texture that are not discriminative are unreliable and should not be considered keypoints. To account for these properties, we use a reliability loss to constrain the score map. We define the reliability of $\boldsymbol{p}_A$ with respect to $\mathcal{I}_B$ based on the matching similarity vector \eqref{equ_sim} as follows:
\begin{equation}
	\label{equ_re}
	r(\boldsymbol{p}_A, \mathcal{I}_B) = \operatorname{softmax}(\operatorname{sim}(\boldsymbol{d}_A, \boldsymbol{D}_B)/t_{rel}),
\end{equation}
where $t_{rel}$ is the temperature for the softmax function. The reliable loss for score map $\boldsymbol{S}_A$ with respect to $\mathcal{I}_B$ is then defined as follows:
\begin{equation}
	\label{equ_Lre}
	\mathfrak{L}_{re}(\boldsymbol{S}_A, \mathcal{I}_B) = \frac{1}{\hat{S}_A}\sum_{\substack{\boldsymbol{p}_A \in \boldsymbol{P}_A, \\ s_A=\boldsymbol{S}_A(\boldsymbol{p}_A)}} (1- r(\boldsymbol{p}_A, \mathcal{I}_B)) * s_A,
\end{equation}
where $\hat{S}_A$ represents the sum of the scores at keypoints. Unlike ALIKE \cite{alike}, we only constrain ${s}_A$ with $r(\boldsymbol{p}_A, \mathcal{I}_B)$ because it only models the reliability of $\boldsymbol{p}_A$. In this formulation, the division of the sum scores $\hat{S}_A$ normalizes the sum of the weighted scores. To obtain a lower loss value, scores with lower weight $(1- r(\boldsymbol{p}_A, \mathcal{I}B))$ should have a higher value, which means that the score $s_A$ should be high when the reliability is higher, and vice versa. The $\mathfrak{L}{re}(\boldsymbol{S}_B, \mathcal{I}A)$ can be obtained in the same way. The $\mathfrak{L}{re}(\boldsymbol{S}_A, \mathcal{I}B)$ and $\mathfrak{L}{re}(\boldsymbol{S}_B, \mathcal{I}_A)$ summarize all the reliability losses of the keypoints $\boldsymbol{p}_A$ and $\boldsymbol{p}B$, respectively. Therefore, the overall reliability loss $\mathfrak{L}{re}$ is the average of the reliability losses of all keypoints.

\subsection{Overall Loss}

The overall loss function for training ALIKED is defined as a weighted sum of the four loss functions described above:
\begin{equation}
	\label{equ_L}
	\mathfrak{L} = \omega_{rp} \mathfrak{L}_{rp} + \omega_{pk} \mathfrak{L}_{pk} + \omega_{ds} \mathfrak{L}_{ds} + \omega_{re} \mathfrak{L}_{re},
\end{equation}
where $\omega{rp}$, $\omega_{pk}$, $\omega_{ds}$, and $\omega_{re}$ are the weights used to balance the losses. During network training, minimizing the dispersity peak loss is straightforward because it involves only a one-dimensional score map. However, minimizing the matching loss is more challenging because it involves high-dimensional descriptors. Therefore, we set $\omega_{pk}=0.5$ and $\omega_{ds}=5$ to compensate for these losses during the training process. The remaining two weights are both set to one during training.

\section{Experiments}
\label{sec_exp}

In this section, we compare the proposed method with SOTA methods that are widely used in visual measurement tasks, including image matching, 3D reconstruction, and visual relocalization. We also perform ablation studies and analyze the limitations of the proposed network.

\subsection{Implementation Details}
By adjusting the number of channels ($c_i$), three networks with different computation costs are designed, as shown in Table. \ref{tab_size}. We take the ALIKE-N \cite{alike} as the baseline network because it has a good balance between running time and matching performance. The radius of the DKD module is two pixels, and keypoints with reprojection distances of less than five pixels are considered ground truth keypoint pairs during training. The normalization temperatures are $t_{det}=0.1$,  $t_{des}=0.1$, and $t_{rel}=1$, which are carefully tuned for different tasks to ensure that the normalized distribution is neither too flat nor too sharp. We use the Adam optimizer \cite{adam} with betas of 0.9 and 0.999 to train the networks. To train the score map, the top 400 keypoints are detected with the DKD, and another 400 more points are randomly sampled. To avoid repeating keypoints in the same area, we apply the NMS to these keypoints and construct the loss function using the remaining points. In the training, the images are resized to $800\times 800$, the batch size is 2, and the gradients are accumulated in 6 batches. We use perspective and homographic datasets together to train the proposed network:

\begin{itemize}
	\item The MegaDepth dataset \cite{megadepth} is used to train perspective image pairs. This dataset collects tourist photos of famous landmarks and uses COLMAP \cite{colmap} to reconstruct the depth and pose for each image. We use the sampled image pairs in the DISK \cite{disk}, which excludes the scenes from the IMW2020 validation and test sets \cite{imw2020}. This dataset has a total of 135 scenes, with 10k image pairs for each scene.
	\item The R2D2 dataset \cite{r2d2} is also used to train homographic image pairs. We use the synthetic image pairs on the Oxford and Paris retrieval datasets \cite{radenovic2018revisiting} and the Aachen dataset \cite{aachen} as well as the synthetic style transferred image pairs on the Aachen dataset\cite{aachen}.
\end{itemize}

We use ALIKED-[N/T](M) to denote the proposed normal/tiny (Table \ref{tab_size}) network with M sample locations (and K=3). We train three networks, namely, the tiny ALIKED-T(16) for real-time performance, the ALIKED-N(16) for the best balance between running time and matching performance, and the ALIKED-N(32) for better matching performance. We train these networks for 100K steps and select the best models based on their matching performance on the validation dataset. 

\subsection{Comparisons with the state-of-the-arts}

To evaluate the performance of the proposed method, we utilize the Intel i7-10700F CPU and NVIDIA GeForce RTX 2060 GPU, along with CUDA 10.2 and pytorch 1.11.0 as the software tools. We compare ALIKED with the following SOTA keypoint and descriptor extraction networks:
\begin{itemize}
    \item \textbf{D2-Net}\cite{d2net}: a network that simultaneously performs description and detection from dense feature maps.
    \item \textbf{LF-Net}\cite{lfnet}: a network that is trained with the virtual target in a two-branch setup.
    \item \textbf{SuperPoint}\cite{superpoint}: a lightweight network that is trained with the Homographic Adaptation strategy.
    \item \textbf{R2D2}\cite{r2d2}: a network that jointly learns the repeatability and reliability maps for keypoint detection.
    \item \textbf{ASLFeat}\cite{aslfeat}: a network that improves the localization accuracy and geometric invariance of D2-Net \cite{d2net}.
    \item \textbf{DISK}\cite{disk}: a method that trains the keypoint and descriptor extraction network with reinforcement learning.
    \item \textbf{ALIKE}\cite{alike}: a lightweight network with a differentiable keypoint detection module.
\end{itemize}

\subsubsection{Real-time Performance}
As shown in Table \ref{tab_hpatches}, ALIKED-T(16) has only 0.192M parameters. To assess the computational complexity, we measure the GFLOPs of different methods on $640\times 480$ images. Due to the sparse descriptor extraction strategy, the ALIKED networks have the lowest GFLOPs compared to existing methods. To compare the running speed, we test the frame rate on $640\times 480$ images with 1K keypoints. Although ALIKED-N(16) has lower GFLOPs than ALIKE-N \cite{alike}, its frame rate is 77.40FPS, which is slightly lower than that of ALIKE-N (84.96FPS), due to the fact that the gathering of image patches is not a standard operation and is not fully optimized (it takes about 1 ms in our implementation). This non-computational operation can be further optimized using low-level techniques. Nevertheless, ALIKED-T(16) achieves a running speed of 125.87 FPS with a matching and reconstruction performance comparable to existing methods (see below).

\begin{table}[!t]
	\centering
	\caption{The matching performance on the Hpatches \cite{hpatches} dataset. ``MP'' denotes million numbers of parameters, GFLOPs, and frames per second (FPS) are evaluated using $640\times 480$ images with 1K keypoints. The top three best results are marked as \textcolor[rgb]{ 1,  0,  0}{\uuline{\textbf{red}}}, \textcolor[rgb]{ 0,  .62,  0}{\uline{\textbf{green}}}, and \textcolor[rgb]{0,0,1}{\uwave{\textbf{blue}}}.}
	\setlength{\tabcolsep}{0.5mm}
	{
 \begin{threeparttable}
	\begin{tabular*}{\linewidth}{@{}@{\extracolsep{\fill}}l|rrrrr@{}}		
    \hline
\multicolumn{1}{c|}{\textit{\textbf{Models}}} & \multicolumn{1}{c}{\textit{\textbf{MP}}} & \multicolumn{1}{c}{\textit{\textbf{GFLOPs}}} & \multicolumn{1}{c}{\textit{\textbf{FPS}}} & \multicolumn{1}{c}{\textit{\textbf{MMA@3}}} & \multicolumn{1}{c}{\textit{\textbf{MHA@3}}} \bigstrut\\
\hline
\textbf{D2-Net(MS)} \cite{d2net} & 7.635  & 889.40 & 7.63   & 37.29\% & 38.33\% \bigstrut[t]\\
\textbf{LF-Net(MS)} \cite{lfnet} & 2.642  & 24.37  & 23.67* & 55.60\% & 57.78\% \\
\textbf{SuperPoint} \cite{superpoint} & 1.301  & 26.11  & 52.63  & 65.37\% & 70.19\% \\
\textbf{R2D2(MS)} \cite{r2d2} & \textcolor[rgb]{ 0,  0,  1}{\uwave{\textbf{0.484}}} & 464.55 & 4.10   & \textcolor[rgb]{ 0,  .62,  0}{\textbf{\uline{75.77\%}}} & 71.48\% \\
\textbf{ASLFeat(MS)} \cite{aslfeat} & 0.823  & 44.24  & 7.10*  & 72.44\% & 73.52\% \\
\textbf{DISK} \cite{disk} & 1.092  & 98.97  & 11.81  & \textcolor[rgb]{ 1,  0,  0}{\textbf{\uuline{77.59\%}}} & 70.56\% \\
\textbf{ALIKE-N} \cite{alike} & \textcolor[rgb]{ 0,  .62,  0}{\textbf{\uline{0.318}}} & 7.91   & \textcolor[rgb]{ 0,  .62,  0}{\textbf{\uline{84.96}}} & 70.78\% & 75.74\% \\
\textbf{ALIKE-L} \cite{alike} & 0.653  & 19.68  & 56.66  & 70.50\% & \textcolor[rgb]{ 0,  0,  1}{\uwave{\textbf{76.85\%}}} \bigstrut[b]\\
\hline
\textbf{ALIKED-T(16)} & \textcolor[rgb]{ 1,  0,  0}{\textbf{\uuline{0.192}}} & \textcolor[rgb]{ 1,  0,  0}{\textbf{\uuline{1.37}}} & \textcolor[rgb]{ 1,  0,  0}{\textbf{\uuline{125.87}}} & 72.99\% & \textcolor[rgb]{ 1,  0,  0}{\textbf{\uuline{78.70\%}}} \bigstrut[t]\\
\textbf{ALIKED-N(16)} & 0.677  & \textcolor[rgb]{ 0,  .62,  0}{\textbf{\uline{4.05}}} & \textcolor[rgb]{ 0,  0,  1}{\uwave{\textbf{77.40}}} & 74.43\% & \textcolor[rgb]{ 0,  .62,  0}{\textbf{\uline{77.22\%}}} \\
\textbf{ALIKED-N(32)} & 0.980  & \textcolor[rgb]{ 0,  0,  1}{\uwave{\textbf{4.62}}} & 75.64  & \textcolor[rgb]{ 0,  0,  1}{\uwave{\textbf{75.23\%}}} & 74.44\% \bigstrut[b]\\
\hline
	\end{tabular*}%
    \begin{tablenotes}
     \item[*] Due to environmental issues, the FPS of LF-Net(MS) \cite{lfnet} and ASLFeat(MS) \cite{aslfeat} are estimated based on the performance ratio between different GPUs.
    \end{tablenotes}
  \end{threeparttable}
	}
	\label{tab_hpatches}%
\end{table}%

\begin{table*}[htbp]
	\centering
	\caption{Stereo matching and multiview reconstruction results on the IMW test set \cite{imw2020} (up to 2048 keypoints). ``NF'', ``Rep'', ``PPC'', ``NM'', ``NL'', and ``TL'' denote the number of features, repeatability, performance per cost,  number of matches, number of landmarks, and track length, respectively. The top three best results are marked as \textcolor[rgb]{ 1,  0,  0}{\uuline{\textbf{red}}}, \textcolor[rgb]{ 0,  .62,  0}{\uline{\textbf{green}}}, and \textcolor[rgb]{0,0,1}{\uwave{\textbf{blue}}}.}
	\begin{threeparttable}
		\setlength{\tabcolsep}{0.9mm}
		{
			\begin{tabular*}{\textwidth}{@{}@{\extracolsep{\fill}}l|r|rrrrrr|rrrrrr@{}}
				\hline
				\hline
				\multicolumn{1}{c|}{\multirow{2}[2]{*}{\textit{\textbf{Methods}}}} & \multicolumn{1}{c|}{\multirow{2}[2]{*}{\textit{\textbf{GFLOPs}}}} & \multicolumn{6}{c|}{\textit{\textbf{Stereo}}}       & \multicolumn{6}{c}{\textit{\textbf{Multiview}}} \bigstrut[t]\\
				&        & \multicolumn{1}{c}{\textit{\textbf{NF}}} & \multicolumn{1}{c}{\textit{\textbf{Rep}}} & \multicolumn{1}{c}{\textit{\textbf{MS}}} & \multicolumn{1}{c}{\textit{\textbf{mAA(5\degree)}}} & \multicolumn{1}{c}{\textit{\textbf{mAA(10\degree)}}} & \multicolumn{1}{c|}{\textit{\textbf{PPC}}} & \multicolumn{1}{c}{\textit{\textbf{NM}}} & \multicolumn{1}{c}{\textit{\textbf{NL}}} & \multicolumn{1}{c}{\textit{\textbf{TL}}} & \multicolumn{1}{c}{\textit{\textbf{mAA(5\degree)}}} & \multicolumn{1}{c}{\textit{\textbf{mAA(10\degree)}}} & \multicolumn{1}{c}{\textit{\textbf{PPC}}} \bigstrut[b]\\
				\hline
				\textbf{D2-Net(MS)} \cite{d2net} & 889.40 & 2045.6 & 16.80\% & 29.30\% & 6.06\% & 12.27\% & 0.01   & 2045.6 & 1999.4 & 3.01   & 17.77\% & 28.30\% & 0.03 \bigstrut[t]\\
				\textbf{SuperPoint} \cite{superpoint} & 26.11  & 2048.0 & 36.40\% & 63.00\% & 19.71\% & 28.97\% & 1.11   & 2048.0 & 1185.4 & 4.33   & 44.35\% & 54.66\% & 2.09 \\
				\textbf{R2D2(MS)} \cite{r2d2} & 464.55 & 2048.0 & 42.90\% & 74.60\% & 27.20\% & 39.02\% & 0.08   & 2048.0 & 1225.9 & 4.28   & 53.13\% & 64.03\% & 0.14 \\
				\textbf{ASLFeat(MS)} \cite{aslfeat} & 77.58  & 2042.6 & 43.10\% & 74.90\% & 22.62\% & 33.65\% & 0.43   & 157.5  & 1106.6 & 4.42   & 45.28\% & 55.61\% & 0.72 \\
				\textbf{DISK} \cite{disk} & 98.97  & 2048.0 & \textcolor[rgb]{ 0,  0,  1}{\uwave{\textbf{44.80\%}}} & \textcolor[rgb]{ 0,  0,  1}{\uwave{\textbf{85.20\%}}} & \textcolor[rgb]{ 0,  0,  1}{\uwave{\textbf{38.72\%}}} & \textcolor[rgb]{ 0,  0,  1}{\uwave{\textbf{51.22\%}}} & 0.52   & 526.4  & 2424.8 & \textcolor[rgb]{ 0,  0,  1}{\uwave{\textbf{5.50}}} & \textcolor[rgb]{ 1,  0,  0}{\textbf{\uuline{63.25\%}}} & \textcolor[rgb]{ 1,  0,  0}{\textbf{\uuline{72.96\%}}} & 0.74 \\
				\textbf{ALIKE-N} \cite{alike} & 7.91   & 1802.7 & 43.30\% & 81.10\% & 35.12\% & 47.18\% & 5.97   & 276.5  & 1644.2 & 4.97   & 59.18\% & 69.21\% & 8.75 \\
				\textbf{ALIKE-L} \cite{alike} & 19.68  & 1770.6 & 42.90\% & 82.20\% & 37.24\% & 49.58\% & 2.52   & 298.3  & 1693.3 & 5.02   & 60.30\% & 70.22\% & 3.57 \bigstrut[b]\\
				\hline
				\textbf{ALIKED-T(16)} & \textcolor[rgb]{ 1,  0,  0}{\textbf{\uuline{1.37}}} & 2031.3 & 43.38\% & 84.01\% & 37.95\% & 50.38\% & \textcolor[rgb]{ 1,  0,  0}{\textbf{\uuline{36.77}}} & 334.9  & 1830.6 & 5.24   & 60.31\% & 70.88\% & \textcolor[rgb]{ 1,  0,  0}{\textbf{\uuline{51.74}}} \bigstrut[t]\\
				\textbf{ALIKED-N(16)} & \textcolor[rgb]{ 0,  .62,  0}{\textbf{\uline{4.05}}} & 1934.2 & \textcolor[rgb]{ 1,  0,  0}{\textbf{\uuline{46.30\%}}} & \textcolor[rgb]{ 0,  .62,  0}{\textbf{\uline{85.47\%}}} & \textcolor[rgb]{ 1,  0,  0}{\textbf{\uuline{39.53\%}}} & \textcolor[rgb]{ 1,  0,  0}{\textbf{\uuline{52.28\%}}} & \textcolor[rgb]{ 0,  .62,  0}{\textbf{\uline{12.91}}} & 401.3  & 1975.4 & \textcolor[rgb]{ 0,  .62,  0}{\textbf{\uline{5.57}}} & \textcolor[rgb]{ 0,  0,  1}{\uwave{\textbf{61.44\%}}} & \textcolor[rgb]{ 0,  0,  1}{\uwave{\textbf{71.78\%}}} & \textcolor[rgb]{ 0,  .62,  0}{\textbf{\uline{17.72}}} \\
				\textbf{ALIKED-N(32)} & \textcolor[rgb]{ 0,  0,  1}{\uwave{\textbf{4.62}}} & 1731.1 & \textcolor[rgb]{ 0,  .62,  0}{\textbf{\uline{45.61\%}}} & \textcolor[rgb]{ 1,  0,  0}{\textbf{\uuline{85.90\%}}} & \textcolor[rgb]{ 0,  .62,  0}{\textbf{\uline{39.52\%}}} & \textcolor[rgb]{ 0,  .62,  0}{\textbf{\uline{52.23\%}}} & \textcolor[rgb]{ 0,  0,  1}{\uwave{\textbf{11.30}}} & 389.9  & 1853.2 & \textcolor[rgb]{ 1,  0,  0}{\textbf{\uuline{5.58}}} & \textcolor[rgb]{ 0,  .62,  0}{\textbf{\uline{61.67\%}}} & \textcolor[rgb]{ 0,  .62,  0}{\textbf{\uline{72.09\%}}} & \textcolor[rgb]{ 0,  0,  1}{\uwave{\textbf{15.59}}} \bigstrut[b]\\
				\hline
			\end{tabular*}%
		}
	\end{threeparttable}
	\label{tab_imw2020}%
\end{table*}%

\begin{figure*}[!t]
	\centering
	\includegraphics[]{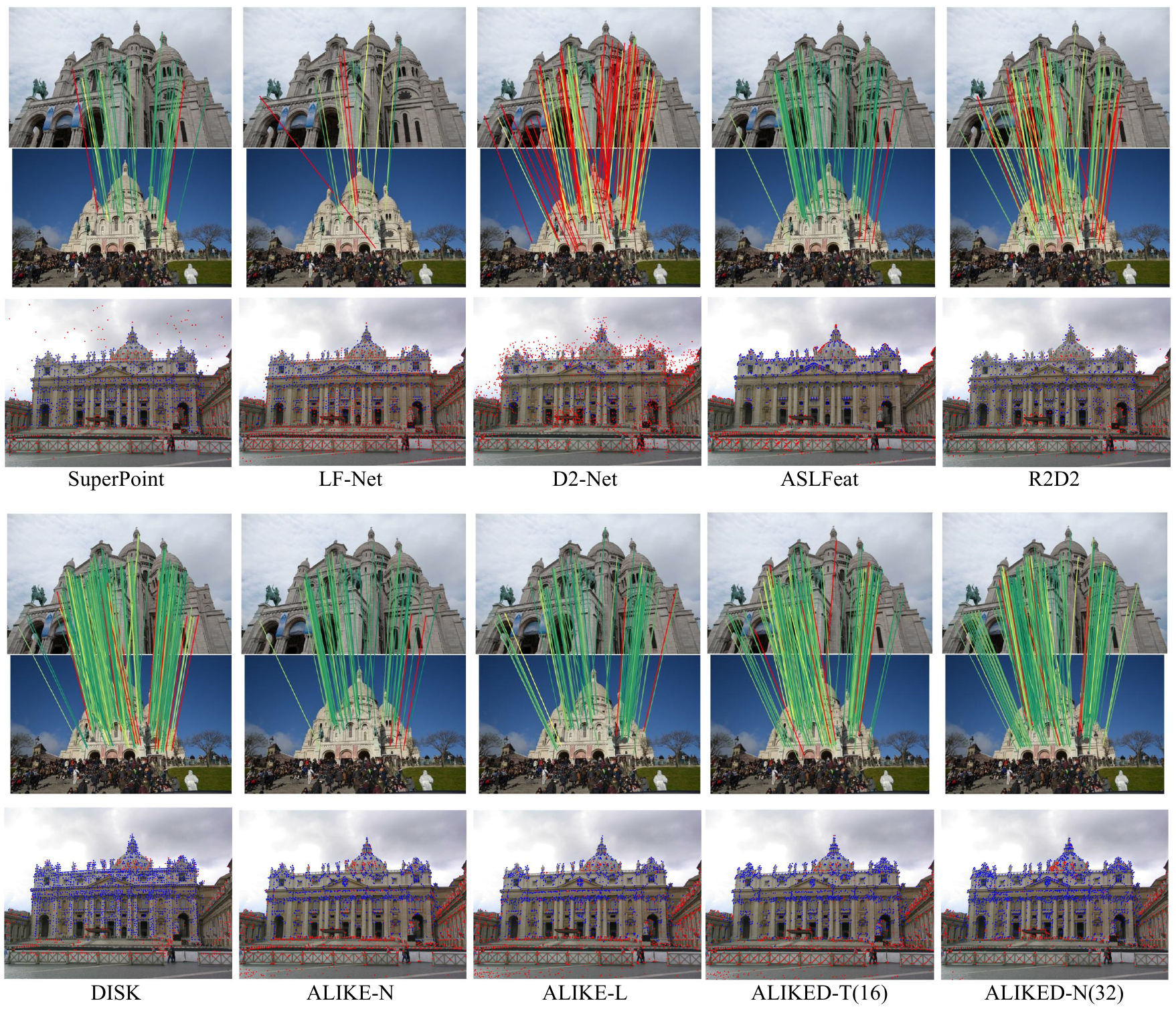}
	\caption{Visualization of the results of stereo matching and multiview reconstruction on IMW-val \cite{imw2020}. For image matching (the first and third rows), the inliers are colored from green to yellow if they are correct (0 to 5 pixels in reprojection error) and red if they are incorrect (more than 5 pixels in reprojection error). For multiview reconstruction (the second and fourth rows), the red and blue points in the image denote the detected and registered keypoints, respectively. Best viewed in color and zoomed in at 400\%.}
	\label{fig_imw}
\end{figure*}

\subsubsection{Homography Image Matching}
We compare the image matching performance of different methods on the Hpatches dataset \cite{hpatches}, which  contains planar homography images of 57 illumination and 79 viewpoint scenes. Each scene contains five image pairs with ground truth homography matrices. According to D2Net \cite{d2net}, eight unreliable scenes are excluded. We extract up to 5000 keypoints with a threshold of 0.2 on the score map and match their descriptors using the mutual Nearest Neighbor (mNN) matcher. Neither the learned matcher \cite{zhao2021probabilistic, superglue} nor the direct image matcher \cite{loftr, cotr, pump} are included, as they are beyond the scope of our study and are not comparable to the proposed method.
Following previous works \cite{superpoint,disk,alike}, the following metrics are evaluated with an error threshold of three pixels (@3 in the following) :
\begin{itemize}
	\item \textbf{MMA} (mean matching accuracy): percentage of correct matches to all estimated putative matches.
	\item \textbf{MHA} (mean homography accuracy): percentage of correct image corners after the image is warped with the estimated homography matrix.
	\item \textbf{MS} (matching score): percentage of correct matches to all co-visible keypoints.
\end{itemize}

Table \ref{tab_hpatches} reports the MMA@3 and MHA@3 on the Hpatches dataset \cite{hpatches}. Since the goal of image matching is to estimate the homography transformation, the MHA is more important than the MMA. Although the DISK \cite{disk} achieves the highest MMA, its MHA is lower than that of ALIKE \cite{alike} and ALIKED. In terms of MHA, the ALIKED-T(16) has the highest value of 78.70\%, followed by ALIKE-N(16) (77.22\%) and ALIKE-L (76.85\%). Despite being a very tiny network, ALIKED-T(16) has the best MHA and a slightly lower MMA than the ALIKE-N, indicating that ALIKED-T(16) is well-balanced in terms of performance and speed.

\begin{table*}[htbp]
	\centering
	\caption{The matching performance on FM-Bench \cite{fmbench}. \%Recall, \%Inlier, and \#Corrs denote the percentage of correct pose estimations, the ratio of inlier matches, and the number of corresponding matches after the RANSAC, respectively. The suffix ``-m'' denotes the metrics before the RANSAC. The top three best results are marked as \textcolor[rgb]{ 1,  0,  0}{\uuline{\textbf{red}}}, \textcolor[rgb]{ 0,  .62,  0}{\uline{\textbf{green}}}, and \textcolor[rgb]{0,0,1}{\uwave{\textbf{blue}}}.}
    \begin{tabular}{l|r|rrrl|rrrl}
\hline
\multicolumn{1}{c|}{\multirow{2}[4]{*}{\textbf{Methods}}} & \multicolumn{1}{c|}{\multirow{2}[4]{*}{\textbf{GFLOPs}}} & \multicolumn{4}{c|}{\textbf{TUM (indoor SLAM settings)}\cite{tumrgbd}} & \multicolumn{4}{c}{\textbf{KITTI (driving settings)}\cite{kitti}} \bigstrut\\
\cline{3-10}       &        & \multicolumn{1}{l}{\textbf{\%Recall}} & \multicolumn{1}{l}{\textbf{\%Inlier}} & \multicolumn{1}{l}{\textbf{\%Inlier-m}} & \textbf{\#Corrs(-m)} & \multicolumn{1}{l}{\textbf{\%Recall}} & \multicolumn{1}{l}{\textbf{\%Inlier}} & \multicolumn{1}{l}{\textbf{\%Inlier-m}} & \textbf{\#Corrs(-m)} \bigstrut\\
\hline
\textbf{LF-Net(MS)} \cite{lfnet} & 24.37  & 53.00  & 70.97  & 56.25  & 143(851) & 80.40  & 95.38  & 84.66  & 202(1045) \bigstrut[t]\\
\textbf{D2-Net(MS)} \cite{d2net} & 889.40 & 34.50  & 67.61  & 49.01  & 74(1279) & 71.40  & 94.26  & 73.25  & 103(1832) \\
\textbf{SuperPoint} \cite{superpoint} & 26.11  & 45.80  & 72.79  & 64.06  & 39(200) & 86.10  & 98.11  & 91.52  & 73(392) \\
\textbf{R2D2(MS)} \cite{r2d2} & 464.55 & 57.70  & 73.70  & 61.53  & 260(1912) & 78.80  & 97.53  & 86.49  & 278(1804) \\
\textbf{ASLFeat(MS)} \cite{aslfeat} & 77.58  & \textcolor[rgb]{ 0,  .69,  .314}{\textbf{\uline{59.90}}} & \textcolor[rgb]{ 0,  .69,  .314}{\textbf{\uline{76.72}}} & \textcolor[rgb]{ 0,  0,  1}{\uwave{\textbf{69.50}}} & 258(1332) & \textcolor[rgb]{ 0,  0,  1}{\uwave{\textbf{92.20}}} & \textcolor[rgb]{ 0,  0,  1}{\uwave{\textbf{98.76}}} & \textcolor[rgb]{ 0,  .69,  .314}{\textbf{\uline{96.16}}} & 630(2222) \\
\textbf{DISK} \cite{disk} & 98.97  & \textcolor[rgb]{ 0,  0,  1}{\uwave{\textbf{59.70}}} & 74.85  & 68.45  & 240(1329) & 90.20  & 98.63  & \textcolor[rgb]{ 1,  0,  0}{\textbf{\uuline{97.34}}} & 506(2527) \\
\textbf{ALIKE-N} \cite{alike} & 7.91   & 45.10  & 73.42  & 67.61  & 60(235) & 89.60  & \textcolor[rgb]{ 0,  .69,  .314}{\textbf{\uline{98.79}}} & 95.76  & 169(687) \\
\textbf{ALIKE-L} \cite{alike} & 19.68  & 43.70  & \textcolor[rgb]{ 1,  0,  0}{\textbf{\uuline{87.49}}} & \textcolor[rgb]{ 1,  0,  0}{\textbf{\uuline{82.82}}} & 81(368) & 88.70  & 98.70  & 96.09  & 193(799) \bigstrut[b]\\
\hline
\textbf{ALIKED-T(16)} & \textcolor[rgb]{ 1,  0,  0}{\textbf{\uuline{1.37}}} & 59.20  & 75.62  & 66.94  & 104(578) & \textcolor[rgb]{ 0,  .69,  .314}{\textbf{\uline{92.30}}} & 98.58  & 95.82  & 409(1405) \bigstrut[t]\\
\textbf{ALIKED-N(16)} & \textcolor[rgb]{ 0,  .62,  0}{\textbf{\uline{4.05}}} & \textcolor[rgb]{ 1,  0,  0}{\textbf{\uuline{63.60}}} & 75.58  & 69.39  & 93(416) & 92.10  & 98.56  & \textcolor[rgb]{ 0,  0,  1}{\uwave{\textbf{96.12}}} & 315(981) \\
\textbf{ALIKED-N(32)} & \textcolor[rgb]{ 0,  0,  1}{\uwave{\textbf{4.62}}} & 58.20  & \textcolor[rgb]{ 0,  0,  1}{\uwave{\textbf{75.91}}} & \textcolor[rgb]{ 0,  .69,  .314}{\textbf{\uline{69.57}}} & 71(301) & \textcolor[rgb]{ 1,  0,  0}{\textbf{\uuline{92.40}}} & \textcolor[rgb]{ 1,  0,  0}{\textbf{\uuline{98.81}}} & 96.06  & 209(643) \bigstrut[b]\\
\hline
\multicolumn{1}{c|}{\multirow{2}[4]{*}{\textbf{Methods}}} & \multicolumn{1}{c|}{\multirow{2}[4]{*}{\textbf{GFLOPs}}} & \multicolumn{4}{c|}{\textbf{T\&T (wide-baseline reconstruction)}\cite{tt}} & \multicolumn{4}{c}{\textbf{CPC (wild reconstruction from web images)}\cite{cpc}} \bigstrut\\
\cline{3-10}       &        & \multicolumn{1}{l}{\textbf{\%Recall}} & \multicolumn{1}{l}{\textbf{\%Inlier}} & \multicolumn{1}{l}{\textbf{\%Inlier-m}} & \textbf{\#Corrs(-m)} & \multicolumn{1}{l}{\textbf{\%Recall}} & \multicolumn{1}{l}{\textbf{\%Inlier}} & \multicolumn{1}{l}{\textbf{\%Inlier-m}} & \textbf{\#Corrs(-m)} \bigstrut\\
\hline
\textbf{LF-Net(MS)} \cite{lfnet} & 24.37  & 57.40  & 66.62  & 60.57  & 54(362) & 19.40  & 44.27  & 44.35  & 50(114) \bigstrut[t]\\
\textbf{D2-Net(MS)} \cite{d2net} & 889.40 & 68.40  & 71.79  & 55.51  & 78(2603) & 31.30  & 56.57  & 49.85  & 84(1435) \\
\textbf{SuperPoint} \cite{superpoint} & 26.11  & 81.80  & 83.87  & 70.89  & 52(535) & 40.50  & 75.28  & 64.68  & 31(225) \\
\textbf{R2D2(MS)} \cite{r2d2} & 464.55 & 73.00  & 80.81  & 65.31  & 84(1462) & 43.00  & 82.40  & 67.28  & 91(954) \\
\textbf{ASLFeat(MS)} \cite{aslfeat} & 77.58  & 88.70  & 85.68  & 79.74  & 327(2465) & 54.40  & 89.33  & 82.76  & 185(1159) \\
\textbf{DISK} \cite{disk} & 98.97  & 86.60  & \textcolor[rgb]{ 0,  .69,  .314}{\textbf{\uline{87.08}}} & \textcolor[rgb]{ 1,  0,  0}{\textbf{\uuline{82.73}}} & 365(3131) & \textcolor[rgb]{ 1,  0,  0}{\textbf{\uuline{59.10}}} & 89.75  & \textcolor[rgb]{ 0,  .69,  .314}{\textbf{\uline{86.30}}} & 310(2266) \\
\textbf{ALIKE-N} \cite{alike} & 7.91   & 82.50  & 83.69  & 76.50  & 92(623) & 40.00  & 87.48  & 82.81  & 80(362) \\
\textbf{ALIKE-L} \cite{alike} & 19.68  & 86.50  & 84.04  & 76.61  & 91(589) & 43.50  & 87.49  & 82.82  & 81(368) \bigstrut[b]\\
\hline
\textbf{ALIKED-T(16)} & \textcolor[rgb]{ 1,  0,  0}{\textbf{\uuline{1.37}}} & \textcolor[rgb]{ 0,  0,  1}{\uwave{\textbf{89.90}}} & 86.19  & 80.87  & 274(1957) & 56.20  & \textcolor[rgb]{ 0,  0,  1}{\uwave{\textbf{90.12}}} & 84.60  & 200(1183) \bigstrut[t]\\
\textbf{ALIKED-N(16)} & \textcolor[rgb]{ 0,  .62,  0}{\textbf{\uline{4.05}}} & \textcolor[rgb]{ 1,  0,  0}{\textbf{\uuline{92.10}}} & \textcolor[rgb]{ 0,  0,  1}{\uwave{\textbf{86.70}}} & \textcolor[rgb]{ 0,  0,  1}{\uwave{\textbf{81.55}}} & 233(1427) & \textcolor[rgb]{ 0,  0,  1}{\uwave{\textbf{58.00}}} & \textcolor[rgb]{ 0,  .69,  .314}{\textbf{\uline{90.68}}} & \textcolor[rgb]{ 0,  0,  1}{\uwave{\textbf{85.99}}} & 183(1035) \\
\textbf{ALIKED-N(32)} & \textcolor[rgb]{ 0,  0,  1}{\uwave{\textbf{4.62}}} & \textcolor[rgb]{ 0,  .69,  .314}{\textbf{\uline{91.70}}} & \textcolor[rgb]{ 1,  0,  0}{\textbf{\uuline{87.49}}} & \textcolor[rgb]{ 0,  .69,  .314}{\textbf{\uline{81.84}}} & 174(1077) & \textcolor[rgb]{ 0,  .69,  .314}{\textbf{\uline{58.30}}} & \textcolor[rgb]{ 1,  0,  0}{\textbf{\uuline{91.37}}} & \textcolor[rgb]{ 1,  0,  0}{\textbf{\uuline{86.75}}} & 153(842) \bigstrut[b]\\
\hline
    \end{tabular}%
	\label{tab_fmbench}%
\end{table*}%

\subsubsection{Pose Estimation and 3D Reconstruction}
We evaluate the performance of stereo pose estimation and 3D reconstruction on the IMW benchmark \cite{imw2020}, where the images were taken by visitors at different times and places with different equipment. Therefore, the appearance of these images varies. We extract up to 2048 keypoints with a threshold of 0.1 on the score map for these images. The benchmark computes the angular difference between the estimated and ground truth translation and rotation vectors and takes the largest of the two as the pose error. The angular error is then thresholded to compute the average accuracy. The mAA(5\degree) and mAA(10\degree) are defined as the mean average accuracy when the angular error is less than 5\degree and 10\degree, respectively. 

Table \ref{tab_imw2020} shows the results of different learning methods on the test set, including repeatability, matching score, mAA, and performance per cost (PPC). The PPC metric is defined as the ratio of mAA(10\degree) to GFLOPs and is used to assess the tradeoff between computational cost and performance of each method. 
For the stereo matching task, the ALIKED-N(16) outperforms the current SOTA method DISK \cite{disk} in terms of Rep, mAA(5\degree), and mAA(10\degree) by 1.5\%, 0.81\%, and 1.06\%, respectively. This improvement can be attributed to the ability to model deformable features, resulting in better descriptors. 
As for the multiview 3D reconstruction task, ALIKED-N(32) performs better than most existing methods except DISK \cite{disk}. This is because DISK produces more matches (NM) than ALIKED, which provides additional constraints in the bundle adjustment process and thus allows for better pose optimization results. However, we observe that the performance of ALIKED-N(32) is similar to that of ALIKED-N(16) on both stereo matching and 3D reconstruction tasks, possibly due to the marginal effect of increasing the number of sample locations beyond 16 on overall performance.
Moreover, despite being a very tiny network, ALIKED-T(16) exhibits only slightly worse performance than the best existing methods on both the stereo matching and multiview reconstruction tasks. The PPC of ALIKED-T(16) on these tasks reaches 36.77 and 51.74, respectively, which is about six times higher than that of the best existing method ALIKE-N\cite{alike}.

\begin{figure}[!t]
	\centering
	\includegraphics[]{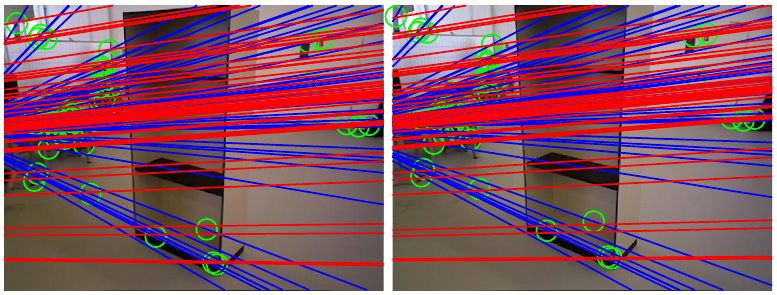}
	\caption{A failure case for image pairs with uneven textures. The green circles represent matched keypoints, the red and blue lines denote the ground truth and estimated epipolar lines.}
	\label{fig_tum}
\end{figure}

For a more intuitive comparison, the matching and reconstruction results of different methods are visualized in Fig. \ref{fig_imw}. SuperPoint \cite{superpoint}, LF-Net \cite{lfnet} and D2-Net \cite{d2net} all have poor matching performance with large viewpoint differences. For R2D2 \cite{r2d2}, the keypoint distribution is scattered, resulting in poorer localization and matches with larger errors (more yellow match lines). The keypoints of DISK \cite{disk} are evenly distributed on the building, resulting in more error-prone matches (yellow match lines). The keypoints of ALIKED inherit the characteristics of ALIKE \cite{alike} and are more concentrated in key regions, such as building edges and corners. Compared to DISK \cite{disk}, ALIKED contains fewer false matches, and compared to ALIKE \cite{alike}, ALIKED recovers more matches. These matches are useful for better matching and reconstruction accuracy.

We also perform comparisons on the FM-Bench \cite{fmbench}, which evaluates the extracted local features on four datasets, namely, the TUM SLAM dataset \cite{tumrgbd}, the KITTI driving dataset \cite{kitti}, the Tanks and Temples (T\&T) dataset \cite{tt}, and the Community Photo Collection (CPC) dataset \cite{cpc}. The  Normalized Symmetric Geometric Distance (NSGD) of the image pairs is evaluated, and its threshold is set to 0.05 by default for the recall calculation. Table \ref{tab_fmbench} reports the results of the evaluation.
In the experiments, we find a typical failure case in the TUM dataset \cite{tumrgbd}, which contains some image pairs with an uneven texture distribution (Fig. \ref{fig_tum}) and many textures on the background. Since the keypoints detected by ALIKED are mainly in texture-rich regions, although the matching results are correct, these matched keypoints are insufficient to establish the geometric constraints, which are not effective for estimating the fundamental matrix. Nevertheless, as shown in Table \ref{tab_fmbench}, ALIKED-N(16) achieves the best recall on the TUM \cite{tumrgbd} and T\&T \cite{tt} datasets, while ALIKED-N(32) achieves the optimal recall on the KITTI dataset \cite{kitti}. Considering all metrics together, ALIEKD-N(32) achieves the best overall matching performance. The matching performance of ALIEKD-T(16) is also comparable to that of existing methods, despite its lower computational requirements.

\begin{table}[!t]
	\centering
	\caption{Visual relocalization results on the Aachen dataset \cite{aachen}. The top three best results are marked as \textcolor[rgb]{ 1,  0,  0}{\uuline{\textbf{red}}}, \textcolor[rgb]{ 0,  .62,  0}{\uline{\textbf{green}}}, and \textcolor[rgb]{0,0,1}{\uwave{\textbf{blue}}}.}
	\setlength{\tabcolsep}{0.5mm}
	{
		\begin{tabular*}{\linewidth}{@{}@{\extracolsep{\fill}}l|rrr|rrr}
\hline
\multicolumn{1}{c|}{\multirow{2}[2]{*}{\textbf{Methods}}} & \multicolumn{3}{c|}{up to 1024 keypoints} & \multicolumn{3}{c}{up to 2048 keypoints} \bigstrut[t]\\
       & \multicolumn{1}{l}{ 0.25m,2\degree} & \multicolumn{1}{l}{0.5m,5\degree} & \multicolumn{1}{l|}{5m,10\degree} & \multicolumn{1}{l}{ 0.25m,2\degree} & \multicolumn{1}{l}{0.5m,5\degree} & \multicolumn{1}{l}{5m,10\degree} \bigstrut[b]\\
\hline
\textbf{D2-Net(SS)} \cite{d2net} & 64.3   & 78.6   & \textcolor[rgb]{ 0,  0,  1}{\uwave{\textbf{91.8}}} & 74.5   & \textcolor[rgb]{ 0,  .62,  0}{\textbf{\uline{85.7}}} & 96.9 \bigstrut[t]\\
\textbf{D2-Net(MS)} \cite{d2net} & 53.1   & 74.5   & 86.7   & 61.2   & 81.6   & 94.9 \\
\textbf{SEKD(SS)} \cite{sekd} & 30.6   & 33.7   & 38.8   & 42.9   & 51.0   & 57.1 \\
\textbf{SEKD(MS)} \cite{sekd} & 35.7   & 42.9   & 50.0   & 50.0   & 63.3   & 70.4 \\
\textbf{SuperPoint} \cite{superpoint} & 58.2   & 66.3   & 72.4   & 69.4   & 78.6   & 87.8 \\
\textbf{R2D2(MS)} \cite{r2d2} & 55.1   & 70.4   & 77.6   & 63.3   & 78.6   & 87.8 \\
\textbf{ASLFeat(SS)} \cite{aslfeat} & 35.7   & 43.9   & 50.0   & 54.1   & 67.3   & 76.5 \\
\textbf{ASLFeat(MS)} \cite{aslfeat} & 25.5   & 32.7   & 41.8   & 49.0   & 59.2   & 69.4 \\
\textbf{DISK} \cite{disk} & 60.2   & 72.4   & 81.6   & 70.4   & 82.7   & 94.9 \\
\textbf{ALIKE-N} \cite{alike} & 59.2   & 73.5   & 83.7   & 68.4   & \textcolor[rgb]{ 0,  0,  1}{\uwave{\textbf{84.7}}} & 96.9 \\
\textbf{ALIKE-L} \cite{alike} & 66.3   & 76.5   & 86.7   & 74.5   & \textcolor[rgb]{ 1,  0,  0}{\textbf{\uuline{87.8}}} & \textcolor[rgb]{ 0,  0,  1}{\uwave{\textbf{98.0}}} \bigstrut[b]\\
\hline
\textbf{ALIKED-T(16)} & \textcolor[rgb]{ 0,  0,  1}{\uwave{\textbf{70.4}}} & \textcolor[rgb]{ 0,  .62,  0}{\textbf{\uline{87.8}}} & \textcolor[rgb]{ 0,  .62,  0}{\textbf{\uline{98.0}}} & \textcolor[rgb]{ 0,  .62,  0}{\textbf{\uline{78.6}}} & \textcolor[rgb]{ 1,  0,  0}{\textbf{\uuline{87.8}}} & \textcolor[rgb]{ 0,  0,  1}{\uwave{\textbf{98.0}}} \bigstrut[t]\\
\textbf{ALIKED-N(16)} & \textcolor[rgb]{ 0,  .62,  0}{\textbf{\uline{73.5}}} & \textcolor[rgb]{ 0,  0,  1}{\uwave{\textbf{85.7}}} & \textcolor[rgb]{ 0,  .62,  0}{\textbf{\uline{98.0}}} & \textcolor[rgb]{ 1,  0,  0}{\textbf{\uuline{80.6}}} & \textcolor[rgb]{ 1,  0,  0}{\textbf{\uuline{87.8}}} & \textcolor[rgb]{ 0,  .62,  0}{\textbf{\uline{99.0}}} \\
\textbf{ALIKED-N(32)} & \textcolor[rgb]{ 1,  0,  0}{\textbf{\uuline{77.6}}} & \textcolor[rgb]{ 1,  0,  0}{\textbf{\uuline{88.8}}} & \textcolor[rgb]{ 1,  0,  0}{\textbf{\uuline{100.0}}} & \textcolor[rgb]{ 0,  0,  1}{\uwave{\textbf{76.5}}} & \textcolor[rgb]{ 1,  0,  0}{\textbf{\uuline{87.8}}} & \textcolor[rgb]{ 1,  0,  0}{\textbf{\uuline{100.0}}} \bigstrut[b]\\
\hline
		\end{tabular*}%
	}
	\label{tab_aachen}%
  \end{table}%

\subsubsection{Visual (re-)localization}
We test ALIKED on the Aachen Day-Night benchmark \cite{aachen}, where we use the default configuration, extract keypoints with a score threshold of 0.1, and test the relocalization performance with up to 1024 and 2048 keypoints. The benchmark first creates a 3D map using the detected keypoints and descriptors from daytime images, uses the detected keypoints and descriptors from query night images to match the 3D map, and evaluates the percentage of correctly matched images under three error thresholds (\textit{i.e.}, (0.25m, 2\degree)/(0.5m,5\degree)/(5m,10\degree)). As shown in Table \ref{tab_aachen}, ALIKED-N(32) has the best visual relocalization performance with up to 1024 and 2048 keypoints. When using up to 1024 instead of 2048 points, the relocalization performance of ALIKED-N(32) degrades slightly, thereby indicating that the extracted keypoints and descriptors of ALIKED-N are very robust. Furthermore, although the descriptor of ALIKED-T(16) has only 64 dimensions, it still outperforms the other methods, especially when only up to 1024 keypoints are used.

\begin{figure}[!t]
	\centering
	\includegraphics[width=\linewidth]{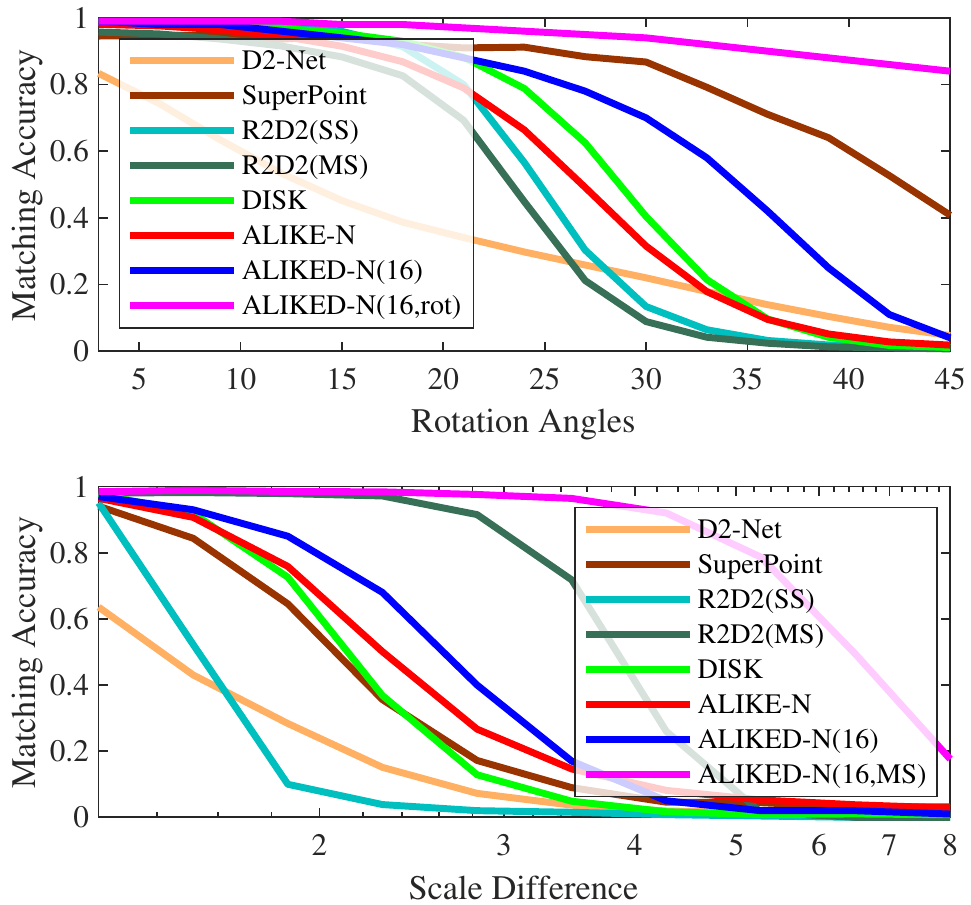}
	\caption{Matching accuracy at different rotation angles (above) and scale differences (below).}
	\label{fig_rs}
\end{figure}

\begin{figure*}[!t]
	\centering
	\includegraphics[width=\linewidth]{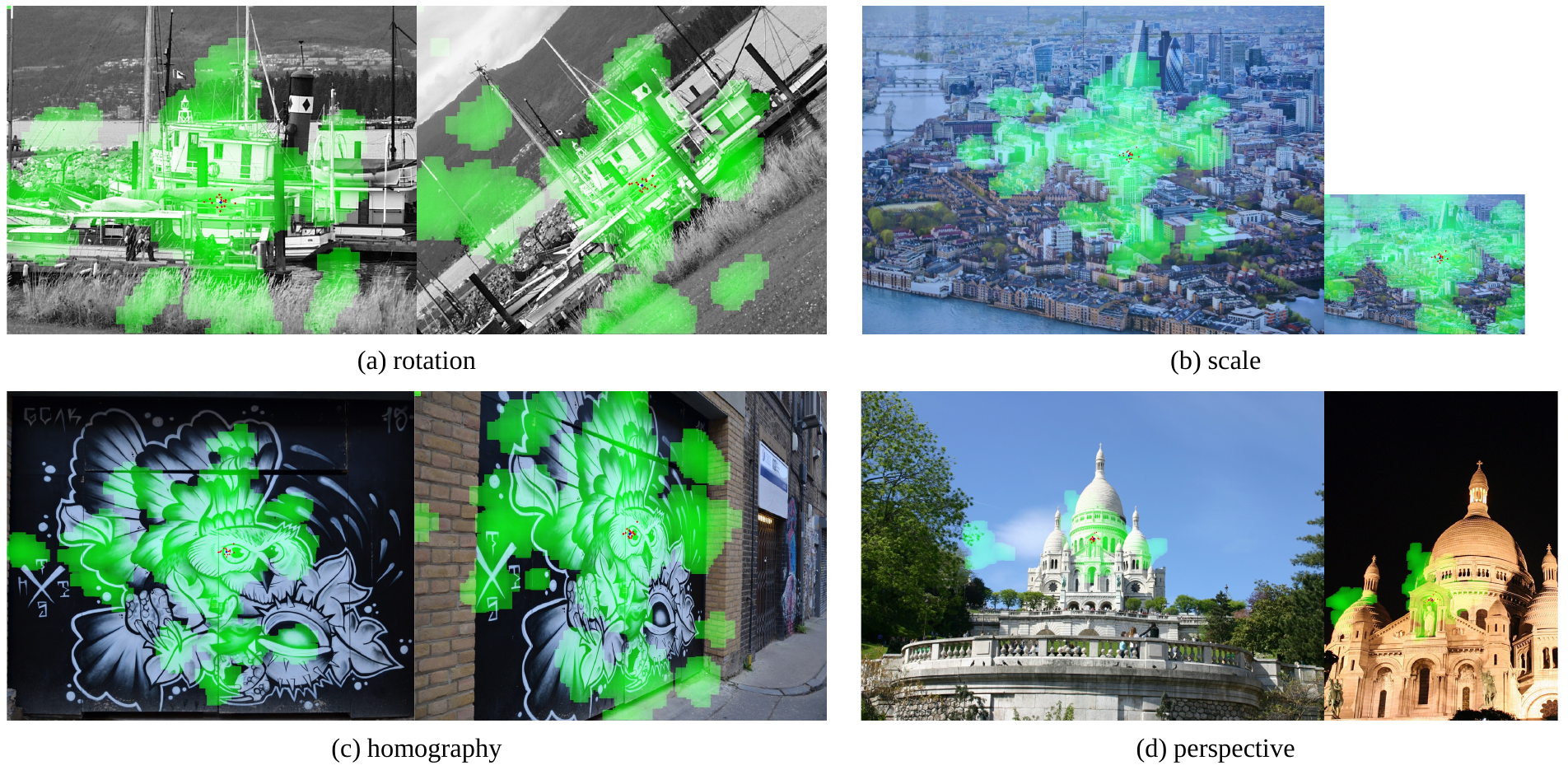}
	\caption{Visualization of the focus areas of deformable descriptors on (a) rotation, (b) scale, (c) homography, and (d) perspective image pairs. The detected keypoints and sample positions of the SDDH are marked as blue and red crosses, respectively, and the respective fields of deformable convolutions on the image are overlaid by green masks. Best viewed in color and zoomed in at 400\%.}
	\label{fig_off}
\end{figure*}

\subsection{Deformable Invariance Evaluation}
To evaluate the deformable invariance of the extracted descriptors, we generate rotation and scaling images based on Hpatches \cite{hpatches}. The reference images in 59 viewpoint scenes are selected and rotated from 0\degree to 45\degree with a step of 3\degree, yielding a total of 840 rotated image pairs. The reference images are scaled by $2^{-s}$, with an $s$ of between 0 and 3 and a step of 0.3, yielding a total of 560 scaled image pairs. We then compare the matching accuracy of ALIKED-N(16) with that of existing methods on these image pairs.

\subsubsection{Rotation Invariance}
As shown in the top chart of Fig. \ref{fig_rs}, ALIKED-N(16, rot) achieves the best rotation invariance, where ALIKED-N(16, rot) is the ALIKED-N(16) trained with rotation image pair augmentation. SuperPoint \cite{superpoint} also achieves good rotation invariance because it was trained with Homography Adaptation, which includes rotation augmentation. Although ALIKED-N(16, rot) has good rotation invariance, it performs slightly worse in 3D reconstruction than ALIKED-N(16), which could be due to the fact that image orientation is an important cue in 3D reconstruction (as is the way humans can sense orientation), and our network implicitly learns to encode this orientation. As a result, rotation augmentation is not used in the training of other networks. Nevertheless, ALIKED-N(16) still outperforms the other methods except for SuperPoint \cite{superpoint} in terms of rotation invariance.

\subsubsection{Scale Invariance}
As shown in the bottom chart of Fig. \ref{fig_rs}, among all single-scale matching methods, ALIKED-N(16) has the best matching accuracy. However, when the scale difference is larger than 4, all single-scale methods degrade to 0, indicating that they cannot handle large-scale differences. For multi-scale (MS) image matching, we use the same multi-scale matching strategy as R2D2(MS) \cite{r2d2}. ALIKED-N(16, MS) degrades when the scale difference is as large as 8, which is much better than R2D2(MS) \cite{r2d2}, because the scaled image usually cannot perfectly match the target scale-difference image in multi-scale matching. As a result, scale invariance at small-scale differences can improve the multi-scale matching performance (and ALIKED-N(16) has a good scale invariance at small-scale differences).

\subsubsection{Visualization of Deformable Descriptors}
We visualize the deformable regions of the corresponding keypoints in four different image pairs in Fig. \ref{fig_off}. The network focuses on the same local structures for the corresponding keypoints in different images. In Fig. \ref{fig_off}(a), the network focuses on the main structure of the ship and rotates along with the image. In Fig. \ref{fig_off}(b), the network does not change the focus area in different images. As a result, the relative receptive field in small-size images is larger than that in large images. Nevertheless, the sample positions of the SDDH change along with the scale, which can provide some scale invariance. Figs. \ref{fig_off}(c) and \ref{fig_off}(d) illustrate the focus areas of the corresponding keypoints in real-world homography and perspective image pairs, respectively. The proposed network can model the focus areas of the corresponding keypoints. 

\subsection{Ablation Studies}
In the ablation studies, we train the networks with different configurations for 10K steps and evaluate the last checkpoint on the Hpatches dataset \cite{hpatches} and IMW-validation set \cite{imw2020}.

\begin{table}[t]
  \centering
  \caption{Ablation studies of NRE loss and training datasets. ``M'' and ``H'' denote the Megadepth \cite{hpatches} and Homographic \cite{imw2020} datasets, respectively. ``NRE'' indicates the type of NRE loss, ``D'' and ``S'' denote dense and sparse NRE losses, respectively. ``Mem'' denotes the GPU memory (GB) when the batch size is 1. The numbers 480 and 800 denote the image resolutions in the training. The metrics are expressed in percentages. The best results are marked as \textbf{bold}.}
  \setlength{\tabcolsep}{0.5mm}
	{
    \begin{tabular*}{\linewidth}{@{}@{\extracolsep{\fill}}cc|c|r|rrr|rr}
    \hline
    \multicolumn{2}{c|}{\textbf{Datasets}} & \multirow{2}[4]{*}{\textbf{NRE}} & \multicolumn{1}{c|}{\multirow{2}[4]{*}{\textbf{Mem}}} & \multicolumn{3}{c|}{\textbf{Hpatches \cite{hpatches}}} & \multicolumn{2}{c}{\textbf{IMW-val \cite{imw2020}}} \bigstrut\\
    \cline{1-2}\cline{5-9}\textbf{M \cite{megadepth}} & \textbf{H \cite{r2d2}} &        &        & \multicolumn{1}{c}{\textbf{MMA@3}} & \multicolumn{1}{c}{\textbf{MHA@3}} & \multicolumn{1}{c|}{\textbf{MS@3}} & \multicolumn{1}{c}{\textbf{mAA(10\degree)}} & \multicolumn{1}{c}{\textbf{MS@3}} \bigstrut\\
    \hline
    480   & -      & D  & 11.1  & 64.87 & 74.26 & 36.56 & \textbf{60.62} & 86.19 \bigstrut\\
    \hline
    480    & -      & \multirow{3}[2]{*}{S} & 3.2   & 63.55 & 72.78 & 32.63 & 56.51 & 83.60 \bigstrut[t]\\
    800    & -      &        & 7.2   & 61.77 & 71.67 & 31.98 & 60.19 & \textbf{86.47} \\
    800    & 800    &        & 9.6   & \textbf{70.72} & \textbf{75.93} & \textbf{44.24} & 54.43 & 83.51 \bigstrut[b]\\
    \hline

    \end{tabular*}%
    }
  \label{tab_abl_data}%
\end{table}%

\begin{table*}[htbp]
	\centering
	\caption{Ablation studies of network configurations. The row numbers before each configuration are quick indexes for reading the main text. ``Stage1'' includes feature extraction and keypoint detection, and ``Stage2'' involves descriptor extraction. The GFLOPs and running times are tested on $640\times 480$ images with 1K keypoints, and the test GPU is a middle-end NVIDIA GeForce RTX 2060. Other abbreviations are defined in Table \ref{tab_abb}. The top three best results are marked as \textcolor[rgb]{ 1,  0,  0}{\uuline{\textbf{red}}}, \textcolor[rgb]{ 0,  .62,  0}{\uline{\textbf{green}}}, and \textcolor[rgb]{0,0,1}{\uwave{\textbf{blue}}}.}
	\setlength{\tabcolsep}{2mm}
	{
    \begin{tabular}{l|cc|rrrr|rrr|rr}
    \hline
    \multicolumn{1}{c|}{\multirow{2}[4]{*}{\textbf{Configurations}}} & \multicolumn{2}{c|}{\textbf{GFLOPs}} & \multicolumn{4}{c|}{\textbf{Running time / ms}} & \multicolumn{3}{c|}{\textbf{Hpatches \cite{hpatches}}} & \multicolumn{2}{c}{\textbf{IMW-val \cite{imw2020}}} \bigstrut\\
    \cline{2-12}       & \multicolumn{1}{c}{\textbf{Stage1}} & \multicolumn{1}{c|}{\textbf{Stage2}} & \multicolumn{1}{c}{\textbf{Extract}} & \multicolumn{1}{c}{\textbf{DKD}} & \multicolumn{1}{c}{\textbf{Desc}} & \multicolumn{1}{c|}{\textbf{Total}} & \multicolumn{1}{c}{\textbf{MMA@3}} & \multicolumn{1}{c}{\textbf{MHA@3}} & \multicolumn{1}{c|}{\textbf{MS@3}} & \multicolumn{1}{c}{\textbf{mAA(10\degree)}} & \multicolumn{1}{c}{\textbf{MS@3}} \bigstrut\\
    \hline
    1: Baseline & \multicolumn{2}{c|}{7.99} & 10.76  & 1.44   & 0.22   & 12.42  & 70.72\% & 75.93\% & 44.24\% & 54.43\% & 83.51\% \bigstrut[t]\\
    2: Baseline+AVG+SELU & \multicolumn{2}{c|}{7.99} & 10.76  & 1.44   & 0.22   & 12.42  & 70.78\% & 75.19\% & 41.28\% & 57.00\% & 82.29\% \bigstrut[b]\\
    \hline
    3: SH1    & \multicolumn{2}{c|}{8.31} & 11.22  & 1.44   & 0.22   & 12.88  & 71.62\% & 74.63\% & 43.40\% & 57.13\% & 83.44\% \bigstrut[t]\\
    4: SH2    & \multicolumn{2}{c|}{9.47} & 12.30  & 1.44   & 0.22   & 13.96  & 71.00\% & \textcolor[rgb]{ 0,  .62,  0}{\textbf{\uline{76.30\%}}} & 44.68\% & 60.87\% & 85.40\% \\
    5: SH3    & \multicolumn{2}{c|}{8.41} & 10.85  & 1.44   & 0.22   & 12.11  & 69.42\% & 75.74\% & 42.69\% & 58.22\% & 84.24\% \bigstrut[b]\\
    \hline
    6: 1xDCN+SH3 & \multicolumn{2}{c|}{8.42} & 10.91  & 1.44   & 0.22   & 12.37  & 71.23\% & 74.81\% & 44.32\% & 60.12\% & 85.93\% \bigstrut[t]\\
    7: 2xDCN+SH3 & \multicolumn{2}{c|}{8.51} & 11.25  & 1.44   & 0.22   & 12.91  & 70.62\% & 75.19\% & 45.50\% & 63.58\% & 87.52\% \bigstrut[b]\\
    \hline
    8: 2xDCN+SH3+SDH1 & 3.48   & 0.13   & 9.60   & 1.44   & 0.66   & 11.70  & 70.63\% & 75.93\% & 45.92\% & 64.30\% & 88.05\% \bigstrut[t]\\
    9: 2xDCN+SH3+SDH2 & 3.48   & 0.59   & 9.60   & 1.44   & 1.25   & 12.29  & 72.49\% & 75.19\% & \textcolor[rgb]{ 0,  0,  1}{\uwave{\textbf{46.07\%}}} & 62.01\% & 86.65\% \\
    10: 2xDCN+SH3+SDH3 & 3.48   & 2.95   & 9.60   & 1.44   & 2.43   & 13.47  & \textcolor[rgb]{ 0,  .62,  0}{\textbf{\uline{72.60\%}}} & \textcolor[rgb]{ 0,  0,  1}{\uwave{\textbf{76.11\%}}} & 45.97\% & 63.68\% & 88.10\% \bigstrut[b]\\
    \hline
    11: 2xDCN+SH3+SDDH1\_16 & 3.48   & 0.54   & 9.60   & 1.44   & 1.61   & 12.65  & 71.53\% & 74.26\% & 44.81\% & \textcolor[rgb]{ 0,  0,  1}{\uwave{\textbf{65.68\%}}} & 89.21\% \bigstrut[t]\\
    12: 2xDCN+SH3+SDDH3\_16 & 3.48   & 0.57   & 9.60   & 1.44   & 1.88   & 12.92  & \textcolor[rgb]{ 0,  0,  1}{\uwave{\textbf{72.52\%}}} & \textcolor[rgb]{ 1,  0,  0}{\textbf{\uuline{76.85\%}}} & \textcolor[rgb]{ 0,  .62,  0}{\textbf{\uline{46.62\%}}} & 65.39\% & 88.93\% \\
    13: 2xDCN+SH3+SDDH5\_16 & 3.48   & 0.64   & 9.60   & 1.44   & 2.19   & 13.23  & 71.94\% & 75.00\% & 44.39\% & 65.51\% & \textcolor[rgb]{ 0,  0,  1}{\uwave{\textbf{89.49\%}}} \bigstrut[b]\\
    \hline
    14: 2xDCN+SH3+SDDH3\_8 & 3.48   & 0.28   & 9.60   & 1.44   & 1.68   & 12.72  & 71.16\% & 75.37\% & 45.85\% & 64.72\% & 88.28\% \bigstrut[t]\\
    15: 2xDCN+SH3+SDDH3\_24 & 3.48   & 0.86   & 9.60   & 1.44   & 2.03   & 13.07  & 69.86\% & 73.52\% & 44.71\% & \textcolor[rgb]{ 0,  .62,  0}{\textbf{\uline{67.59\%}}} & \textcolor[rgb]{ 1,  0,  0}{\textbf{\uuline{90.29\%}}} \\
    16: 2xDCN+SH3+SDDH3\_32 & 3.48   & 1.14   & 9.60   & 1.44   & 2.18   & 13.22  & \textcolor[rgb]{ 1,  0,  0}{\textbf{\uuline{72.64\%}}} & 75.00\% & \textcolor[rgb]{ 1,  0,  0}{\textbf{\uuline{47.37\%}}} & \textcolor[rgb]{ 1,  0,  0}{\textbf{\uuline{67.78\%}}} & \textcolor[rgb]{ 0,  .62,  0}{\textbf{\uline{90.12\%}}} \bigstrut[b]\\
    \hline
    \end{tabular}%
		}
	\label{tab_abl}%
\end{table*}%

\begin{table}[htbp]
	\centering
	\caption{Abbreviations of network configurations in Table. \ref{tab_abl}. (KxK,dim) denotes the $K\times K$ convolution layer with output feature dimensions of $dim$.}
        \begin{tabular*}{\linewidth}{@{}@{\extracolsep{\fill}}|c|l|}
        \hline
        nxDCN  & Using deformable convolution in the last n blocks. \bigstrut\\
        \hline
        SH1    & Score Head 1: [(3x3,1),sigmoid]. \bigstrut\\
        \hline
        SH2    & \makecell[l]{Score Head 2: [(3x3,d),SELU,[(3x3,4),SELU]x2,\\(3x3,4),sigmoid].} \bigstrut\\
        \hline
        SH3    & \makecell[l]{Score Head 3: [(1x1,8),SELU,[(3x3,4),SELU]x2,\\(3x3,4),sigmoid].} \bigstrut\\
        \hline
        SDH1   & Sparse Descriptor Head 1: [(1x1,d),SELU,(1x1,d)]. \bigstrut\\
        \hline
        SDH2   & Sparse Descriptor Head 2: [(3x3,d)]. \bigstrut\\
        \hline
        SDH3   & Sparse Descriptor Head 3: [(3x3,d),SELU,(3x3,d)]. \bigstrut\\
        \hline
        SDDHK\_M & \makecell[l]{Sparse Deformable Descriptor Head with kernel size K \\and M sample positions.} \bigstrut\\
        \hline
        \end{tabular*}%
	\label{tab_abb}%
\end{table}%

\subsubsection{Ablation Studies on NRE Loss and Training Data}
We train the baseline network with different settings to study the NRE loss and training data. As shown in Table \ref{tab_abl_data}, the dense NRE loss is better than the sparse NRE loss. However, since the SDDH only extracts sparse descriptors, we can only use the sparse NRE loss in the training. Fortunately, the sparse NRE loss uses significantly less GPU memory than the dense NRE loss. To compensate for the performance degradation caused by using the sparse NRE loss, we increase the image resolution from $480\times 480$ to $800\times 800$, which improves the matching performance on the IMW-validation set \cite{imw2020}, as shown in the third row. To further improve the matching performance on homography image pairs, we include the homography dataset \cite{r2d2}, which improves the MMA and MS on Hpatches \cite{hpatches}, but degrades the performance on the IMW-validation \cite{imw2020} due to the fact that the baseline network is not powerful enough to model features for both homography and perspective images.

\subsubsection{Ablation Studies on Network Architecture}
We improve the baseline network from three perspectives:

\textbf{Feature Extraction}:
For accurate visual measurements, the extracted image feature should have a good localization performance and a large respective field on the image for a robust descriptor extraction. We expect to achieve these goals by improving the baseline network ALIKE-N \cite{alike}. As shown in the first two rows of Table \ref{tab_abl}, we start by changing max-pooling to average-pooling (AVG) and ReLU to SELU \cite{selu}. These changes improve the mAA(10\degree) on the IMW-validation \cite{imw2020} by 3.43\% but have limited improvements on the other metrics.
Furthermore, for geometric invariance feature extraction, we replace the vanilla convolutions in the last two blocks with DCN \cite{dcn} as shown in the sixth and seventh rows of Table \ref{tab_abl}. Compared to the network with vanilla convolution (fifth row), the network with DCN \cite{dcn} in the last two blocks (2xDCN) increases the computation by only 0.1 GFLOPs and the running time by only 0.8 ms. Besides improving the MS@3 on the Hpatches \cite{hpatches} by 2.81\%, this change also improves the mAA(10\degree) and MS@3 on the IMW-validation \cite{imw2020}  by 4.36\% and 3.28\%, respectively.
Since the first two blocks have higher feature resolution, using DCN \cite{dcn} would significantly increase the computational cost. Moreover, since the front blocks are responsible for low-level feature extraction, using DCN \cite{dcn} may degrade performance. Therefore, we use DCN \cite{dcn} only  in the last two blocks.

\textbf{Score Head}:
Due to computational cost considerations, the baseline network has a score head of only one $1\times1$ convolutional layer. In this paper, we propose an efficient descriptor extraction pipeline that does not require the extraction of dense descriptor maps, but instead extracts descriptors at keypoint locations, thereby greatly reducing the computational cost. As a result, we can use more complex score heads (SH1-SH3). As shown in Table \ref{tab_abb}, we first increase the size of the convolutional kernel to 3 (SH1). This simple modification improves the MMA@3 and MS@3 on Hpatches \cite{hpatches} by 0.84\% and 2.12\%, respectively, and the MS@3 on IMW-validation \cite{imw2020} by 1.15\%. 
We also design the SH2 with deeper layers. As shown in the ninth and eighth rows of Table \ref{tab_abl}, the SH2 increases the MHA@3 and MS@3 on Hpatches \cite{hpatches} by 1.67\% and 1.28\%, respectively, and the mAA(10\degree) and MS@3 on IMW-validation \cite{imw2020} by 3.71\% and 1.96\%, respectively. However, compared to the baseline, the SH2 increases the GFLOPs and running time by 1.41 and 1.54 ms, respectively. Through careful examination, we find that the first $3\times 3$ convolution contributes the most to computational cost. Therefore, we design SH3, which performs a $1\times 1$ convolution to reduce the feature channels to 8 before estimating the score map. As shown in Table \ref{tab_abl}, SH3 saves 1.06GFLOPs compared to SH2 with similar matching performance.

\textbf{Descriptor Head}:
In the baseline network, the descriptor head is a simple $1\times1$ convolutional layer. 
To improve efficiency, we sample feature patches from the feature map and then use these feature patches to extract sparse descriptors. Table \ref{tab_abb} shows how we develop the sparse descriptor heads SDH1-SDH3 and SDDH. The SDH1 uses two $1\times1$ convolutional layers, the SDH2 uses one $3\times3$ convolutional layer for a larger respective field, and the SDH3 adopts two $3\times3$ convolutional layers. 
For the SDHs, since only the descriptors at the keypoint locations are extracted, the computational cost is proportional to the number of keypoints. To evaluate their efficiency, we split the network into two stages, with the first stage including feature extraction and DKD, and the second stage including descriptor extraction. As shown in Table \ref{tab_abl}, the GFLOPs and running time are tested for each step with 1K keypoints. Deeper and wider descriptor heads result in better overall performance. Specifically, compared to SDH1, SDH3 improves MMA@3 and MHA@3 on Hpatches \cite{hpatches} by 1.97\% and 0.19\%, respectively, and increases mAA(10\degree) and MS@3 on IMW-validation \cite{imw2020} by 1.68\% and 1.45\%, respectively. However, compared to SDH1, SDH3 increases the computational cost by 2.82 GFLOPs and 1.77 ms per 1K keypoints. In addition, SDHs still use conventional convolutions, which do not provide geometric invariance. To address these issues, we propose the SDDH in Section \ref{sec_SDDH}, which provides local geometric invariance through estimated sample positions. To find the appropriate configurations of the SDDH, we vary its kernel size K (1,3,5) and the number of sample locations M (8,16,24,32) as shown in Table \ref{tab_abl}. The kernel size K slightly improves the matching performance, and as M increases, the descriptors have a larger respective field, thereby allowing the network to find more supporting features on the feature map and resulting in more powerful descriptors. Therefore, we identify K=3 and M=16 as the best tradeoff between runtime and performance, and we use a larger network with M=32 for better performance.

\begin{figure}[t]
	\centering
	\includegraphics[]{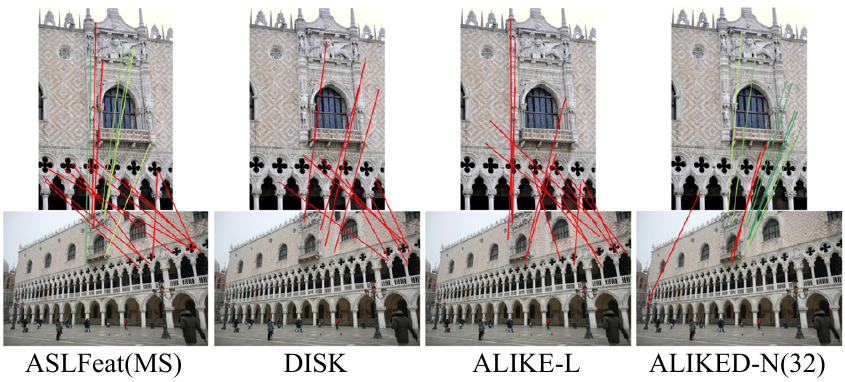}
	\caption{Matching results of SOTA methods for images with large differences in scale and viewpoint. The matches are colored from green to yellow based on their reprojection error, which ranges from 0 to 5 pixels. False matches are marked in red. Best viewed in color and zoomed in at 400\%.}
	\label{fig_limitations}
\end{figure}

\subsection{Limitations of ALIKED}
Although ALIKED performs well in various visual measurement tasks, it still has some limitations. First, for image matching tasks involving significant differences in both scale and viewpoint, ALIKED may have difficulty obtaining correct matches, as shown in Fig. \ref{fig_limitations}. It should be noted, however, that this challenge is not just to ALIKED and is also shared by other SOTA keypoint descriptor methods such as ALSFeat \cite{aslfeat}, DISK \cite{disk}, and ALIKE \cite{alike}. ALSFeat can recover several matches due to its multi-scale matching strategy, while ALIKED can also recover some correct matches due to its deformable descriptor. However, to save computational effort, the SDDH in ALIKED has only one layer for deformable position estimation, so it has limitations in modeling image deformation. Therefore, it may fail in cases where there are both significant scale and viewpoint differences. To overcome this limitation, one possible solution is to use a learning-based matcher \cite{superglue,zhao2021probabilistic} instead of a simple mNN matcher. We chose to use the mNN matcher in our study because we focused on the performance of pure keypoint descriptors. Second, ALIKED uses grid-sampling and produces 32-bit floating descriptors, which may not be ideal for mobile platforms. Therefore, one of our future research goals is to develop a hardware-friendly keypoint descriptor extraction network based on ALIKED.

\section{Conclusions}
\label{sec_con}

In this paper, we propose the SDDH (Sparse Deformable Descriptor Head) and design the ALIKED (A LIghter Keypoint and descriptor Extraction network with deformable transformation). Unlike  existing keypoint and descriptor extraction networks, the proposed method incorporates a deformable transformation into the descriptors, making them more robust. Moreover, the SDDH extracts descriptors only on sparse keypoints, which eliminates redundant convolutions in the descriptor map extraction and reduces the running time. To train the sparse deformable descriptors extracted obtained from the proposed network, we further relax the neural reprojection error loss from dense to sparse. The experiments demonstrate that the proposed network achieves excellent performance in important visual measurement tasks, including image matching, 3D reconstruction, and visual relocalization. In our future work, we plan to train the keypoint descriptor extraction and matching networks simultaneously, develop hardware-friendly keypoint descriptor networks, and further improve the performance of the network.

\normalem

\end{document}